\documentclass[preprint,12pt,authoryear]{elsarticle}

\input{psfig.sty}
\usepackage{epsfig}
\usepackage{amsthm} 
\usepackage{amssymb}
\usepackage{multirow}

\newtheorem{mydef}{Definition}

\journal{Neural Networks}

\begin{document} 

\begin{frontmatter}

\title{ OCReP: An Optimally Conditioned Regularization for Pseudoinversion 
Based Neural Training\\}

\author[DipInf]{ Rossella Cancelliere } 
\author[OATo]{ Mario Gai } 
\author[LIP6]{ Patrick Gallinari }
\author[DipInf]{ Luca Rubini}

\address[DipInf]{University of Turin, Dep. of Computer Sciences, 
C.so Svizzera 185, 10149 Torino, Italy 
}

\address[OATo]{National Institute of Astrophysics, Astrophys. 
Observ. of Torino, Pino T.se (TO), Italy }

\address[LIP6]{Laboratory of Computer Sciences, LIP6, Univ. Pierre 
et Marie Curie, Paris, France }

\begin{abstract}

In this paper we consider the training of single hidden layer 
neural networks by pseudoinversion, which, in spite of its 
popularity, is sometimes affected by numerical instability issues. 
Regularization is known to be effective in such cases, so that 
we introduce, in the framework of Tikhonov regularization, a 
matricial reformulation of the problem which allows us 
to use the condition number as a diagnostic tool for 
identification of instability. 
By imposing well-conditioning requirements on the relevant matrices, 
our theoretical analysis allows the identification of an optimal 
value for the regularization parameter from the standpoint of 
stability. 
We compare with the value derived by cross-validation for 
overfitting control and optimisation of the generalization 
performance. 
We test our method for both regression and classification tasks. 
The proposed method is quite effective in terms of predictivity, 
often with some improvement on performance with respect to 
the reference cases considered. 
This approach, due to analytical determination of the regularization 
parameter, dramatically reduces the computational load required by 
many other techniques. 
\end{abstract}


\end{frontmatter}

\section{Introduction}
\label{intro}

In past decades Single Layer Feedforward Neural Networks (SLFN) training 
was mainly accomplished by iterative algorithms involving the repetition 
of learning steps aimed at minimising the error functional over 
the space of network parameters. These techniques often gave rise to methods slow 
and computationally expensive. 

Researchers therefore have always been motivated to explore alternative algorithms 
and recently some new techniques based on matrix inversion have been 
developed. 
In the literature, they were initially employed to train radial basis 
function neural networks \citep{poggio1}: the idea of using 
them also for different neural architectures was suggested for instance 
in \citep{cancelliere}.

The work by Huang et al. (see for instance \citep{huang}) gave rise to a 
great interest in neural network community: they presented the technique of Extreme Learning Machine 
(ELM) for which SLFNs with randomly chosen input weights and 
hidden layer biases can learn sets of observations with a desired 
precision, provided that activation functions in the hidden layer are 
infinitely differentiable. 
Besides, because of the use of linear output neurons, output weights determination 
can be brought back to linear systems solution, obtained via Moore-Penrose generalised inverse 
(or pseudoinverse) of the hidden layer output matrix; so doing iterative training 
is no more required. 

Such techniques appear anyway to require more hidden units with respect 
to conventional neural network training algorithms 
to achieve comparable accuracy, as discussed in Yu and Deng \citep{yu}. 

Many application-oriented studies in the last years have been devoted to the use 
of these single-pass techniques, easy to implement and computationally fast; 
some are described e.g. 
in \citep{nguyen,kohno,ajorloo}. 
A yearly conference is currently being held on the subject, the International 
Conference on Extreme Learning Machines, and the method is currently 
dealt with in some journal special issue, e.g. Soft Computing \citep{Soft} and 
the International Journal of Uncertainty, Fuzziness and Knowledge-Based 
Systems \citep{Uncert}. 

Because of the possible presence of singular and almost singular matrices, 
pseudoinversion is known to be a powerful but numerically unstable method: 
nonetheless in the neural network community it is often used without singularity checks and evaluated through approximated methods.

In this paper we improve on the theoretical framework using singular 
value analysis to detect the occurrence of instability. 
Building on Tikhonov regularization, which is known to be effective in this 
context \citep{GolubHansen}, we present a technique, named Optimally 
Conditioned Regularization for Pseudoinversion (OCReP), that 
replaces unstable, ill-posed problems with well-posed ones. 

Our approach is based on the formal definition of a new matricial formulation 
that allows the use of condition number as diagnostic tool. 
In this context an optimal value for the regularization parameter is analytically 
derived by imposing well-conditioning requirements on the relevant matrices. 

The issue of regularization parameter choice has often been identified as crucial in literature, and dealt with in a number of historical contributions: a conservative guess might 
put its published estimates at several dozens. Some of the most relevant works
are mentioned in section \ref{OLSRR}, where the related theoretical background is recalled.

Its determination, mainly aimed at overfitting control, has often been done either experimentally via cross-validation, requiring heavy computational training procedures, or analytically under 
specific conditions on the matrices involved, sometimes hardly applicable to real datasets, as discussed in section \ref{OLSRR}. 

In section \ref{pseudo} we present the basic concepts concerning input and output 
weights setting, 
and we recall the main ideas on ill-posedness, regularization and condition number. 

In section \ref{theory} our matricial framework is introduced, 
and constraints on condition number are imposed in order to 
derive the optimal value for the regularization parameter. 

In section \ref{res} our diagnosis and control tool is tested on some applications 
selected from the UCI database and validated by comparison with the framework regularized via cross-validation and with the unregularized one. 

The same datasets are used in section \ref{comparison} to test the technique effectiveness: our performance is compared with those obtained in other regularized frameworks, originated in both statistical and neural domains.

\section{ Recap on ordinary least-square and ridge regression estimators }
\label{OLSRR}

As stated in the introduction, pseudoinversion based neural training brings back output weights determination to linear systems solution: in this section we recall some general ideas on this issue, that in next sections will be specialized to deal with SLFN training.

The estimate of $ \beta $ through ordinary least-squares (OLS) technique is a classical tool for solving the problem 
\begin{equation}
Y = X \beta + \epsilon \ , 
\label{eq:ORL}
\end{equation}
where $Y$ and $\epsilon$ are column $n$-vectors, $\beta$ is a 
column $p$-vector and $X$ is an $n \times p$ matrix; $\epsilon$ 
is random, with expectation value zero and variance $\sigma^2$. 

In \citep{Hoerl62} and \citep{HoerlKennard70} the role of ordinary ridge regression 
(ORR) estimator $\hat{\beta}(\lambda)$ as 
an alternative to the OLS estimator in the presence of multicollinearity is deeply analized. 
In statistics, multicollinearity (also collinearity) is a phenomenon in which two or more 
predictor variables in a multiple regression model are highly correlated, meaning that 
one can be linearly predicted from the others with a non-trivial degree of accuracy. 
In this situation, the coefficient estimates of the multiple regression may change erratically 
in response to small changes in the model or the data. 

It is known in literature that there exist estimates of $\beta$ with smaller mean square error 
(MSE) than the unbiased, or Gauss-Markov, estimate  \citep{golubwahba,berger} 
\begin{equation}
\hat{\beta}(0) = (X^T X)^{-1} X^T Y \ . 
\label{eq:hatbeta}
\end{equation}

Allowing for some bias may result in a significant variance reduction: this is known as the bias-variance dilemma (see e.g. \citep {tibshirani,geman}, whose effects on output weights determination will be deepened in section \ref{stabgen}.

Hereafter we focus on  
the one parameter family of ridge estimates 
$\hat{\beta}(\lambda)$ given by 
\begin{equation}
\hat{\beta}(\lambda) = (X^T X + n \lambda I)^{-1} X^T Y \ . 
\label{eq:hatbetalambda}
\end{equation}

It can be shown that $\hat{\beta}(\lambda)$  is also the solution to 
the problem of finding the minimum over $\beta$ of 
\begin{equation}
{1 \over n} || Y - X \beta ||^2_2 + \lambda ||\beta||^2_2 \ , 
\label{eq:lagrangian}
\end{equation}
which is known as 
the method of regularization in the approximation theory literature \citep{golubwahba}; basing on it we will develop the theoretical framework for our work in the next sections.

There has always been a substantial amount of interest in estimating a good value of $\lambda$ from the 
data: in addition to those already cited in this section a non-exhaustive list of well known or more recent papers is e.g \citep{HoerlKennard2,lawless,mcdonald, nordberg, saleh, kibria, khalafshukur, Mardikyan}. 

A meaningful review of these formulations is provided in \citep{DorugadeKashid}. They first define the matrix $T$ such that  $T^TX^T X T = \Lambda $ ($ \Lambda = diag (\lambda_1, \lambda_2, \cdots \lambda_p ) $ contains the eigen values of the matrix $ X^T X $ ); then they set $Z = X T$ and $\alpha = T^T \beta$, and show that a great amount of different methods require the OLS estimates of $\alpha$ and $\sigma$  

\begin{equation}
\hat{\alpha} = (Z^T Z)^{-1} Z^T Y \ , 
\label{eq:alphas}
\end{equation}

\begin{equation}
\hat{\sigma}^2 = { Y^T Y - \hat{\alpha}^T Z^T Y \over  n - p - 1} \ .
\label{eq:variance}
\end{equation}

\noindent to define effective ridge parameter values. It is important 
to note  that often specific conditions on data are needed to evaluate 
these estimators.
 
In particular this applies to the expressions of the ridge parameter 
proposed by \citep{kibria} and \citep{HoerlKennard70}, that share the 
characteristic of being functions of the ratio between $\hat{\sigma}^2$ 
and a function of $\hat{\alpha}$; they will be used for comparison with 
our proposed method in section \ref{comparison}. 

The alternative technique of generalised cross-validation (GCV) proposed by \citep{golubwahba}
provides a good estimate of $\lambda$ from the data as the minimizer of 
\begin{equation}
V(\lambda) = 
{1 \over n} 
{ || I - A(\lambda) Y||^2_2
\over 
\left[ {1 \over n} {\rm Trace} (I - A(\lambda)) \right]^2_2} \ , 
\label{eq:Golub1}
\end{equation}
where 
\begin{equation}
A(\lambda) = X (X^T X + n \lambda I)^{-1} X^T \ . 
\label{eq:Golub2}
\end{equation}

This solution is particularly interesting, since it does not require an estimate of $\sigma^2$: because of this, it will be one term of comparison with our experimental results in section \ref{comparison}.


In the next section we will show how the problem of finding a good solution to (\ref{eq:ORL}) applies to the context of pseudoinversion based neural training, specializing the involved relevant matricies to deal with this issue.

\section{Main ideas on regularization and condition number theory  }
\label{pseudo}

\subsection{ Generalised inverse matrix for weights setting  }
\label{UnregPseudoInv}
We deal with a standard SLFN with $L$ input neurons, $ M$ hidden neurons and $Q$ output neurons, non-linear activation functions $ \phi $ in the hidden layer and linear activation functions in the output layer.

Considering a dataset of $N$ distinct training samples 
$ (\textbf {x}_j, \textbf {t}_j)$, where 
$ \textbf {x}_j \in  \mathbb{R}^L $ and $ \textbf {t}_j \in  \mathbb{R}^Q $, 
the learning process for a SLFN aims at producing the matrix of desired 
outputs  $T  \in  \mathbb{R}^{N \times Q} $ 
when the matrix of all input instances $X  \in  \mathbb{R}^{N \times L}$ is presented as input.

As stated in the introduction, in the pseudoinverse approach the 
matrix of input weights and hidden layer biases is randomly 
chosen and no longer modified: we name it $ C $. 
After having fixed $ C $, the hidden layer output matrix $ H=\phi (XC) $ 
is completely determined; we underline that since 
$H \in  \mathbb{R}^{N \times M} $, it is not invertible. 

The use of {\it linear} output neurons allows to determine the output 
weight matrix $W^*$ in terms of the OLS solution to the problem 
$ T = H \, W \, + \epsilon$, in analogy with eq.(\ref{eq:ORL}). 
Therefore from eq.(\ref{eq:hatbeta}), we have 
\begin{equation}
W^*=(H^T H)^{-1} H^T T \  
\label{eq:5}
\end{equation}

According to \citep{penrose,bishop} 
\begin{equation}
W^*=H^+ T \, . 
\label{eq:5b}
\end{equation}

$ H^+ $ is the Moore-Penrose pseudoinverse (or generalized inverse) of matrix $ H $, and it minimises the cost functional 
\begin{equation}
E_D =  || H W-T||_2^2 \, 
\label{eq:4}
\end{equation}
\noindent

Singular value decomposition (SVD) is a computationally simple and accurate 
way to compute the pseudoinverse (see for instance \citep{golub}), 
as follows.

Every matrix $H \in  \mathbb{R}^{N \times M} $ can be expressed as 
\begin{equation}
H=U \Sigma V^T \, ,
\label{eq:6}
\end{equation}
where $ U \in  \mathbb{R}^{N \times N}$  and $ V \in  \mathbb{R}^{M \times M} $ are 
orthogonal matrices and $ \Sigma \in  \mathbb{R}^{N \times M} $ is a 
rectangular diagonal matrix (i.e. a matrix with $ \sigma_{ih}=0 $ if $ i \neq h$); 
its elements $ \sigma _{ii} = \sigma _i $, called 
singular values, are non-negative. 
A common convention is to list the singular values in descending order, i.e. 
\begin{equation}
\sigma_1 \geq \sigma_2 \geq \cdots \geq \sigma_p > 0  \; \; \; \; \; \; \; \; \; \; \; \;  
\label{eq:sing_values}
\end{equation}
\noindent 
where p = min  $\{ N,M \}$,  so that $ \Sigma $ is uniquely determined.

The SVD of $ H $ is then used to obtain the pseudoinverse matrix $ H^+ $:
 
\begin{equation}
H^+ =V \Sigma^+ U^T, 
\label{eq:pseudoinverse}
\end{equation}
where $\Sigma^+ \in  \mathbb{R}^{M \times N}$ is again a rectangular diagonal matrix whose elements $\sigma^+_i$ 
are obtained by taking the reciprocal of each corresponding 
element: $\sigma^+ _i = 1 / \sigma _i $ (see also \citep{rao}). 
From eq.(\ref{eq:5}) we than have:

\begin{equation}
W^*=V \Sigma^+U^T T, 
\label{eq:sol_unreg}
\end{equation}

\noindent 
{\bf Remark}
\\ 
An interesting case occurs when only $k < p$ elements in 
eq.(\ref{eq:sing_values}) 
are non-zero, i.e. $\sigma_{k+1} = \cdots = \sigma_p = 0 $; in this case the rank of matrix  $H$ is $ k $ and $\Sigma^+ $ is defined as:
\begin{equation}
\Sigma^+=\textrm{diag} (1 / \sigma_1, \cdots, 1 / \sigma_k, 0, \cdots, 0) \;\; \in \mathbb{R}^{M \times N},
\end{equation}
\noindent as shown for instance in \citep{golub}.

This is also often done in practice, for computational reasons, for elements 
smaller than a predefined threshold, thus actually computing an approximated 
version of the pseudoinverse matrix $H^+$. 

This approach is for example used by default for pseudoinverse evaluation 
by means of the Matlab \textbf{pinv} function 
\footnote{http://www.mathworks.com/help/matlab/ref/pinv.html.},  
because the tool is widely used by many scientists for example in ELM context, each time that it is applied blindly, i.e. without having decided at what threshold to zero the small $ \sigma _i $, an approximation {\it a priori} uncontrolled is introduced in $H^+$ evaluation.

\subsection{Stability and generalization properties of regularization algorithms }
\label{stabgen}

A key property for any learning algorithm is stability:  the learned mapping has to suffer only small changes in presence of small perturbations (for instance the deletion of one example in the training set).

Another important property is generalization: the performance on the training examples (empirical error) 
must be a good indicator of the performance on future examples 
(expected error), that is, the difference between the two must be small. 
An algorithm that guarantees good generalization predicts well if its empirical error is small.

Many studies in literature dealt with the connection between stability and generalization: the notion of stability has been investigated by several authors, e.g. by Devroye and Wagner \citep{Devroye} and Kearns and Ron \citep{kearns}. 

Poggio et al. in  \citep{Mukherjee1} introduced a statistical form of leave-one-out stability, named $CVEEE_{loo}$, building on a cross-validation leave-one-out stability endowed with conditions on stability of both expected and empirical errors; they demonstrated that this condition is necessary and sufficient for generalization and consistency of the class of empirical risk minimization (ERM) learning algorithms, and that it is also a sufficient condition for generalisation for not ERM algorithms (see also \citep{Poggio3}). 

To turn an original instable, ill-posed problem into a well-posed one, regularization methods of the form (\ref{eq:lagrangian}) 
are often used \citep{badeva} and among them, Tikhonov regularization is one of the most common 
\citep{tikhonov1,tikhonov2}. It minimises the error functional
\begin{equation}
E\equiv E_D + E_R=|| H W-T||_2^2 + ||\Gamma W ||_2^2,
\label{eq:reg1}
\end{equation}

\noindent obtained adding to the cost functional $ E_D $ in eq.(\ref{eq:4}) a penalty term $E_R$ that depends on a 
suitably chosen Tikhonov matrix $\Gamma $. 
This issue has been discussed in its applications to neural networks in 
\citep{poggio}, and surveyed in \citep{girosi,haykin}. 

Besides, Bousquet and Elisseeff \citep{Bousquet} proposed the notion of uniform 
stability to characterize the generalization properties of an 
algorithm. Their results state that Tikhonov regularization algorithms are 
uniformly stable and that uniform stability implies good generalization  \citep{Mukherjee2}. 

Regularization thus introduces a penalty function that not only improves 
on stability, making the problem less sensitive to initial conditions, 
but it is also important to contain model complexity avoiding overfitting.

The idea of penalizing by a square function of weights is also well known in neural literature as weight decay: a wide amount of articles have been devoted to this argument, and more generally to the advantage of regularization for the control of overfitting.  Among them we recall \citep{hastie,tibshirani,bishop,girosi,fu,gallinari}.

A frequent choice is $ \Gamma = \sqrt \gamma I $, 
to give preference to solutions with smaller norm \citep{bishop}, so eq. (\ref{eq:reg1}) 
can be rewritten as
\begin{equation}
E\equiv E_D + E_R=|| H W-T||_2^2 + \gamma || W ||_2^2. 
\label{eq:reg3}
\end{equation}


We define $ \hat{W} = \min_{W} (E) $ the regularized solution of (\ref{eq:reg3}): it belongs to the family of ridge estimates described by eq.(\ref{eq:hatbetalambda}) and can be expressed  as 
\begin{equation}
\hat{W} =(H^T H + \gamma I)^{-1} H^T T \,
\label{minimum_reg}
\end{equation}
or, as shohw in (\citep{fuhry}) as

\begin{equation}
\hat{W} = V D U^T T.
\label{eq:tikmat}
\end{equation}

$ V $ and $ U $ are from the singular value decomposition of $ H $ (eq.(\ref{eq:6})) 
and $ D  \in  \mathbb{R}^{M \times N}$ is a rectangular diagonal matrix whose elements, 
built using the singular values $ \sigma_i $ of 
matrix $ \Sigma$, are:
\begin{equation}
 D_{i} = \frac{\sigma _i}{\sigma _i ^2 + \gamma}.
\label{eq:tiksing}
\end{equation}

We remark on the difference between the minima of the regularized 
and unregularized error functionals. 
Increasing values of the regularization parameter $\gamma$ induce larger 
and larger departure of the former (eq.~(\ref{minimum_reg})) from 
the latter (eq.~(\ref{eq:5})). 
Thus, the regularization process increases the bias of the 
approximating solution and reduces its variance, as discussed about 
the bias-variance dilemma in section \ref{OLSRR}.

A suitable value for the Tikhonov parameter $\gamma$ 
has therefore to derive from a compromise between 
having it sufficiently large to control the approaching 
to zero of $\sigma_i$ in eq.(\ref{eq:tiksing}), 
while avoiding an excess of the 
penalty term in eq.(\ref{eq:reg3}). 
Its tuning is therefore crucial. 

\subsection{Condition number as a measure of ill-posedness}
\label{regul}
The condition number of a matrix $ A \in  \mathbb{R}^{N \times M} $ is 
the number $ \mu (A) $ defined as
\begin{equation}
\mu (A)= ||A || \, ||A^+ ||
\label{eq:numcond}
\end{equation}
\noindent 
where $ \left\| \cdot \right\| $ is any matrix norm. 
If the columns (rows) of $A$ are linearly independent, e.g. in case of 
experimental data matrices, then $ A^+ $ is a left (right) inverse of $A$, 
i.e.    $ A^+ A = I_N $ $ (A A^+ = I_M) $. 
The Cauchy-Schwarz inequality in this case then provides 
$ \mu (A) \geq 1 $; 
besides, $ \mu (A) \equiv \mu(A^+)$ . 

Matrices are said to be ill-conditioned if $ \mu (A) \gg 1 $.

If $ \left\| \cdot \right\|_2 $ norm is used, then 
\begin{equation}
\mu (A) = \frac{\sigma_{1}(A)}{\sigma_{p}(A)} ,
\label{eq:ncond}
\end{equation}

\noindent where $ \sigma_{1} $ and $ \sigma_{p} $ are the largest 
and smallest singular values of $ A $ respectively.

From eq.(\ref{eq:ncond}) we can easily understand that large condition numbers 
$ \mu (A) $ suggest the presence of very small singular values (i.e. of almost 
singular matrices), whose numerical inversion, required to evaluate $ \Sigma^+$ 
and the unregularized solution $ W^* $, is a cause of instability. 

From numeric linear algebra we also know that  if the condition number is large 
the problem of finding least-squares solutions to the corresponding system of linear 
equations is ill-posed, i.e. even a small perturbation in the data can lead to huge 
perturbations in the entries of solution (see \citep{golub}). 

According to \citep{Mukherjee2} the stability of Tikhonov regularization algorithms can also be characterized using the classical notion of condition number: our proposed regularization method fits within this context.
We will see that it specifically aims at analitically determining the value of the $\gamma$ parameter that minimizes the conditioning of the regularized hidden layer output matrix so that the solution $ \hat{W} $ is stable in the sense of eq.(2.9) of  \citep{Mukherjee2}.

In the next section, we will derive the optimal value of the regularization 
parameter $\gamma$ according to this stability criterion (minimum condition 
number). 

The experimental results presented in sections \ref{res} and \ref{comparison} will evidence that our quest for stable solutions allows us to also achieve good generalization and predictivity. A comparison will be made to this purpose with the performance obtained when $ \gamma$ is determined via the standard cross-validation approach, aimed at overfitting control and generalization performance optimization.

\section{Conditioning of the regularized matricial framework} 
\label{theory}

For convenient implementation of our diagnostics, 
and building on eq.(\ref{eq:tikmat}), 
we propose an original matricial framework in which to develop 
our study tool with the following definition. 

\begin{mydef}
We define the matrix 
\begin{equation}
H^{reg} \equiv V D U^T
\label{eq:reg_matrix}
\end{equation}
\noindent as the {\bf regularized} hidden layer output matrix of the 
neural network.
\end{mydef}

\noindent 
This allows us to rewrite eq.(\ref{eq:tikmat}) as 
\begin{equation}
\hat{W} =  H^{reg} T \ , 
\label{eq:Reg_Sys}
\end{equation}
for similarity with eq.(\ref{eq:5}). 

By construction, $H^{reg}$ is decomposed in three matrices according to the SVD framework, 
and its singular values are provided by eq.(\ref{eq:tiksing}) 
as a function of the singular values $\sigma_i$ of $H$. 

This new regularized matricial framework
makes easier the comparison of the properties of $H^{reg}$ with those
of the corresponding unregularized matrix $H^+$. 
In fact, when unregularized pseudoinversion is used, nothing prevents 
the occurrence 
of very small singular values that make numerically instable the evaluation of $ H^+$ (see eq. \ref{eq:pseudoinverse}).
On the contrary, even in presence of very small 
values $\sigma_i$ of the original unregularized problem, a careful choice 
of the parameter $\gamma$ allows to tune the singular values $D_{i}$ of 
the regularized matrix $ H^{reg} $, preventing numerical instability. 

\subsection{Condition number definition}
\label{mureg_def}

According to eq. (\ref{eq:ncond}), we define the condition number of $ H^{reg} $ as:

\begin{equation}
\mu (H^{reg}) = \frac{D_{max}}{D_{min}} \ .
\label{eq:ncond_reg2}
\end{equation}
\noindent where $D_{max} $ and $ D_{min}$ are 
the largest and smallest singular values of $ H^{reg}$.

\begin{figure}
\begin{centering}
\psfig{figure=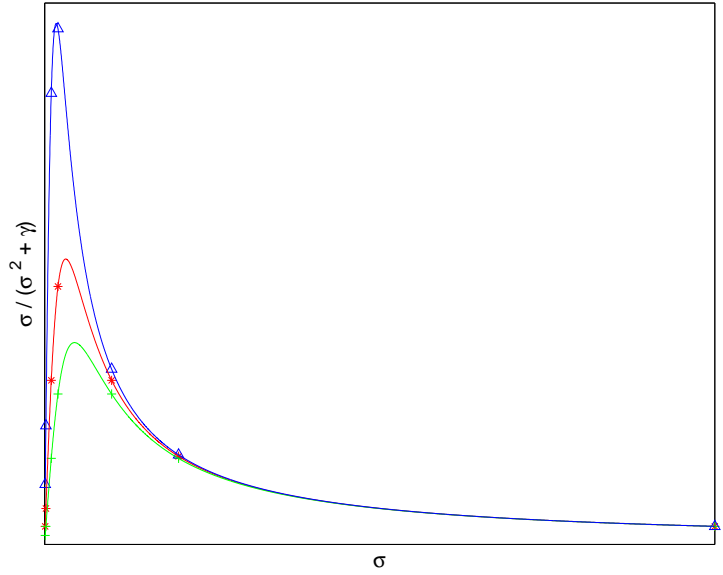,width=0.7\textwidth,height=0.4\textheight} 
\par\end{centering}
\caption{Example of regularized/unregularized singular values relationship via eq.~(\ref{eq:tiksing})}
\label{fig:Dfunction}
\end{figure}

The shape of the functional relation $  \sigma / (\sigma ^2 + \gamma) $ 
that links regularized and unregularized singular values, defined through 
eq. (\ref{eq:tiksing}), is shown in Fig.\ref{fig:Dfunction} for three 
different values of $ \gamma$. 

The curves are non-negative, because $ \sigma>0$ and $ \gamma>0$, and 
have only one maximum, with coordinates 
$ (\sqrt \gamma; \frac{1}{2 \sqrt \gamma}) $. 

A few pairs of corresponding values $ (D_i, \sigma_i) $ are marked by dots 
on each curve.

For the sake of the determination of $ \mu (H^{reg}) $ we are interested in 
evaluating $D_{max} $ and $ D_{min}$ of $H^{reg}$ over the finite, 
discrete range $[\sigma_1, \sigma_2, \ldots, \sigma_p]$. 

The value $D_{max}$ is reached in correspondence to a given singular value 
of $H$, a priori not known, that we label $\sigma_{max}$, so that:

\begin{equation}
D_{max} = \frac{\sigma_{max}}{\sigma_{max}^2 + \gamma}.
\label{eq:maxsingval}
\end{equation}

The variation of $ \gamma$ has the effect of changing the curve 
and shifting its maximum point 
within the interval  $ [\sigma_1 , \sigma_p] $. 
Therefore, $ \sigma_{max} $ can coincide with any singular value of $H$ 
from eq.~(\ref{eq:sing_values}), including the extreme ones. 

Conversely, we now demonstrate that $ D_{min}$ can only be reached in 
correspondence to $ \sigma_1$ or $ \sigma_p$ (or both when coincident).

\bigskip
{\bf Theorem 3.1}

The minimum singular value $ D_{min}$ of matrix $ H^{reg} $ can only be reached 
in correspondence to the largest singular value $ \sigma_1$ or to the smallest 
singular value $ \sigma_p$ of the unregularized matrix $ H $ (or both).
\smallskip

\begin{proof} 
Without loss of generality, we can express $ \gamma$ as a function of 
$ \sigma_1 \sigma_p$, i.e. $\gamma = \beta \sigma_1 \sigma_p$, 
where $\beta$ is a real positive value. 
By replacement in eq.~(\ref{eq:tiksing}), we get 
$$
D_1 = \frac{1}{\sigma_1 + \beta \sigma_p}, \;\;\;\;\;\; 
D_p = \frac{1}{\sigma_p + \beta \sigma_1}
$$

To establish their ordering, we evaluate the difference $ \Delta $ of their 
inverses: 
$$
\Delta = \frac{1}{D_1} - \frac{1}{D_p} = 
(\sigma_1 + \beta \sigma_p)-(\sigma_p + \beta \sigma_1) = 
(1 - \beta) (\sigma_1 - \sigma_p) \ . 
$$

\noindent 

Recalling that $\sigma_1 - \sigma_p > 0$, we can distinguish three cases:

\begin{description}
  \item[Case 1, $\beta > 1 \ (\gamma >  \sigma_1 \sigma_p) \ \rightarrow \ \Delta < 0 \ 
\rightarrow \ D_1 > D_p$] \hfill \\
Because of the $ D_i$ distribution shape, $D_p$ is also the minimum among all values 
$ D_i$, 
so that $ D_{min} \equiv D_p$. 
\\ 
  \item[Case 2, $\beta < 1 \ (\gamma <  \sigma_1 \sigma_p) \ \rightarrow \ \Delta > 0 \ 
\rightarrow \ D_1 < D_p$] \hfill \\
Then, $D_1$ is also the minimum among all values $ D_i$, so that $ D_{min} \equiv D_1$.
\\ 
  \item[Case 3, 
$\beta = 1 \ (\gamma =  \sigma_1 \sigma_p) \ \rightarrow \ \Delta = 0 \ 
\rightarrow \ D_1 = D_p$] \hfill \\
Thus, $D_1$ and $D_p $ are both minima, so that $ D_{min} \equiv D_1 = D_p$.
\end{description}
\end{proof}

\subsection{Condition number evaluation}
\label{mureg_eval}

The result by Theorem 3.1 allows us to find, according to eq. (\ref{eq:ncond_reg2}), 
the following expressions for $ \mu (H^{reg})$ :

\noindent {\bf Case 1, $\beta > 1$:} 
$$ 
\mu (H^{reg}) = \frac{D_{max}}{D_p} = \frac{\sigma_{max}(\sigma_p+ \beta \sigma_1)}{\sigma_{max}^2 + \beta \sigma_1 \sigma_p}
$$
\noindent {\bf Case 2, $\beta < 1$:} 
$$ 
\mu (H^{reg}) = \frac{D_{max}}{D_1} = \frac{\sigma_{max}(\sigma_1+ \beta \sigma_p)}{\sigma_{max}^2 + \beta \sigma_1 \sigma_p}
$$
\noindent {\bf Case 3, $\beta = 1$:} 
$$ 
\mu (H^{reg}) = \frac{D_{max}}{D_p} =  \frac{D_{max}}{D_1}= \frac{\sigma_{max}(\sigma_p+ \sigma_1)}{\sigma_{max}^2 + \sigma_1 \sigma_p}
$$

Bearing in mind that well-conditioned problems are characterized by small condition numbers, we now will look for the $ \beta$ parameter values 
which, in the three cases above, make the regularized condition number 
smaller.

\begin{figure}
\begin{centering}
\psfig{figure=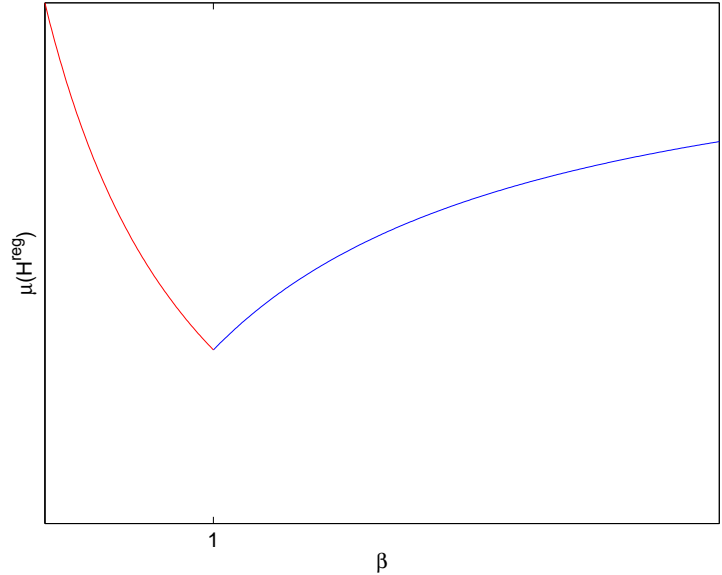,width=0.7\textwidth,height=0.4\textheight} 
\par\end{centering}
\caption{Regularized condition number vs. $ \beta $}
\label{fig:mu_trend}
\end{figure}

In Case 1, $ \mu (H^{reg}) $ is an increasing function of $ \beta $, 
so that in its domain, i.e. $(1, \infty)$, 
its minimum value is reached when $\beta \rightarrow 1^+ $.
On the contrary, in Case 2, $ \mu (H^{reg}) $ is a decreasing function of 
$ \beta$, so that in its domain, i.e. $ (0, 1) $, the minimum is reached 
when $\beta \rightarrow 1^- $. 

Fig.\ref{fig:mu_trend} shows the function behaviour over the whole 
domain. 

Both cases have a common limit: 

\begin{equation}
\lim_{\beta \to 1^+}\mu (H^{reg}) = 
\lim_{\beta \to 1^-}\mu (H^{reg}) = 
\frac{\sigma_{max}(\sigma_p+ \sigma_1)}{\sigma_{max}^2 + \sigma_1 \sigma_p}
\end{equation}
\noindent 
Such value is just that provided by Case 3, which can therefore be considered 
the best possible choice to minimize the condition number. 

Thus our quest for the best possible conditioning for the matrix $ H^{reg}$ 
identifies an explicit optimal value for the regularization 
parameter $\gamma $: 
\begin{equation}
\gamma = \sigma_1 \sigma_p 
\label{eq:opt_par_gamma}
\end{equation}
\noindent

\section{Simulation and Discussion}
\label{res}

For the numerical experimentation, we use eight benchmark datasets 
from the UCI repository \citep{Bache} listed in Table \ref{TableData}. 
All simulations are carried out in Matlab 7.3 environment. 

The performance is assessed by statistics over a set of 50 different 
extractions of input weigths, computing either the average RMSE (for 
regression tasks) or the average percentage of misclassification rate (for classification 
tasks) on the test set. 
Either quantity is labeled ``Err" in the tables summarising our results. 
The error standard deviation (labeled ``Std") is also computed to evidence 
the dispersion of experimental results. 

Our regularization strategy, labeled Optimally Conditioned Regularization for Pseudoinversion (OCReP), is 
verified by simulation against the common approach in which cross-validation is used i) to determine the regularization parameter $ \gamma$ at a fixed high number of hidden neurons and ii) to perform also hidden neurons number optimization, respectively in sec.~\ref{sec:OCReP_CV} and \ref{sec:OCReP_unreg}.

A discussion of the effectiveness of OCReP in terms of minimization of the condition number of the involved matricies is done in sec.~\ref{sec:final_cond}. 

\begin{table}
\begin{tabular}{lllll}
\hline\noalign{\smallskip}
Dataset & Type & N. Instances & N. Attributes & N. Classes \\
\noalign{\smallskip}\hline\noalign{\smallskip}
Abalone & Regression & 4177 & 8 & - \\
Machine Cpu & Regression & 209 & 6 & - \\
Delta Ailerons & Regression & 7129 & 5 & - \\
Housing & Regression & 506 & 13 & - \\
Iris & Classification & 150 & 4 & 3 \\
Diabetes & Classification & 768 & 8 & 2 \\
Wine & Classification & 178 & 13 & 3 \\
Segment & Classification & 2310 & 19 & 7 \\
\noalign{\smallskip}\hline
\end{tabular}
\caption{The UCI datasets and their characteristics }
\label{TableData}       
\end{table}

\subsection {OCReP performance assessment: fixed number of hidden units}
\label{sec:OCReP_CV}

In this section  we compare OCReP with a regularization approach in which $\gamma$ is 
selected by a cross-validation scheme, which is typically used  for control of under/overfitting and optimization of the model  generalization performance. 
A 70\%/30\% split between training and test set is applied; then, 
a three-fold cross-validation search on the training set 
identifies the best $\gamma$ by best performance on the validation 
set, over the set of 50 values of  $\gamma$ $[10^{-25}, 10^{-24}, \cdots 
10^{25}]$.

For the sake of comparison, a fixed, high number of hidden units $M$ 
is used, selected according to dimension and complexity of the datasets. 
For the three datasets Machine Cpu, Iris and Wine the simulation is performed for 50 and 100 hidden neurons; for Abalone, 
Delta Ailerons, Housing and Diabetes, we use 50, 100, 200 and 300 neurons; 
for Segment, we use 1000 and 1500 units. 

\begin{figure}
\centering

\includegraphics[width=0.49\textwidth]{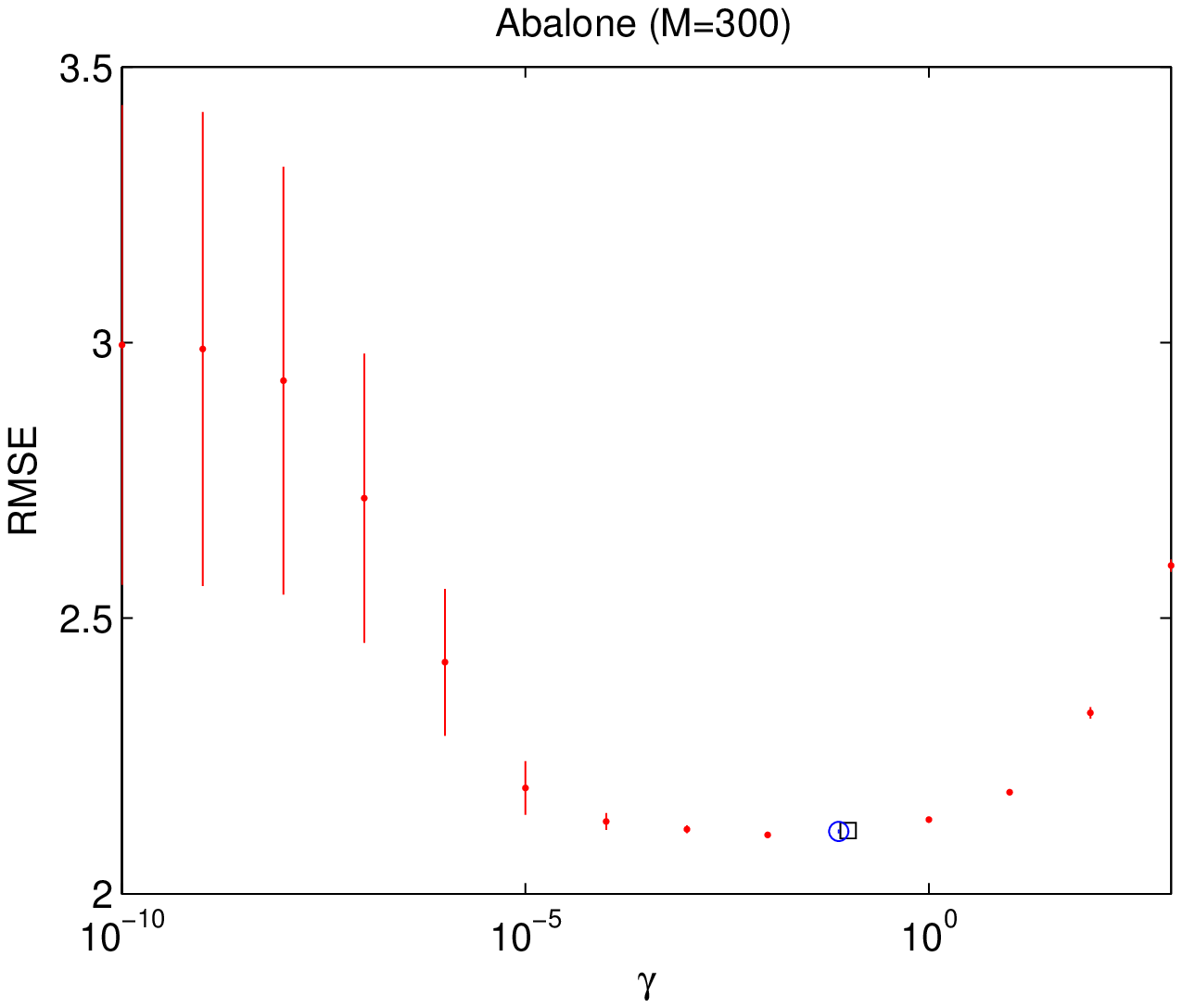}
\includegraphics[width=0.49\textwidth]{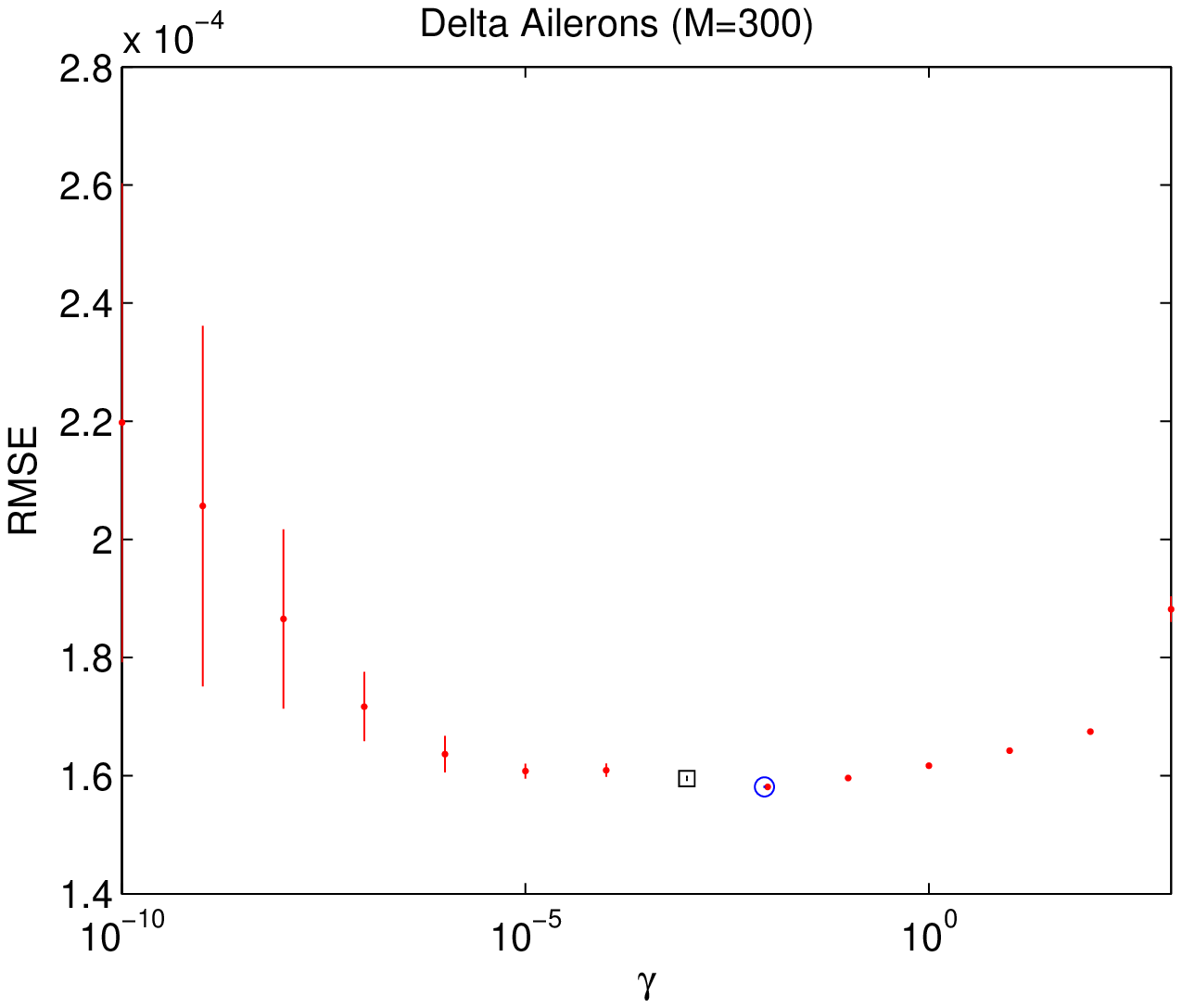}

\includegraphics[width=0.49\textwidth]{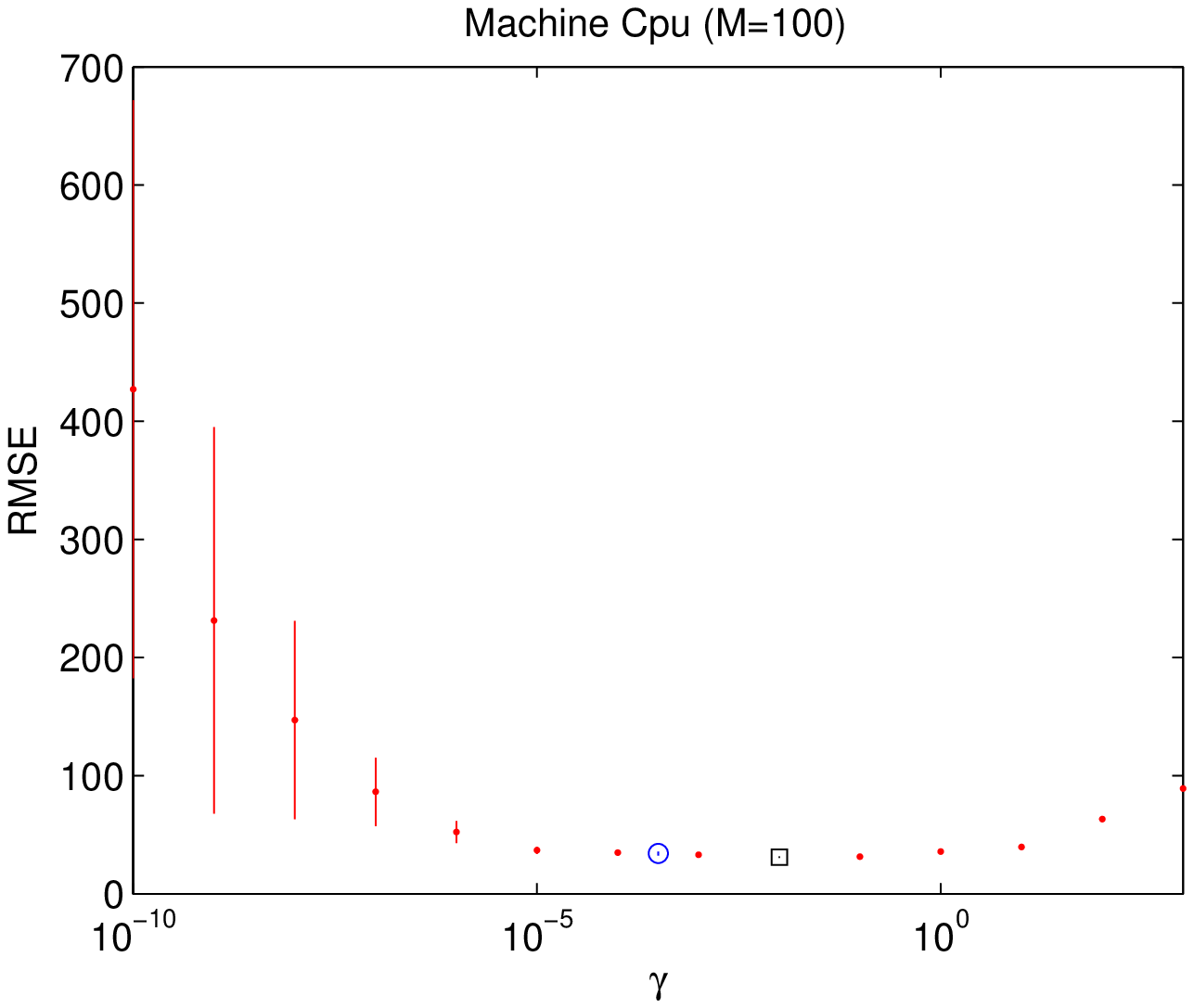}
\includegraphics[width=0.49\textwidth]{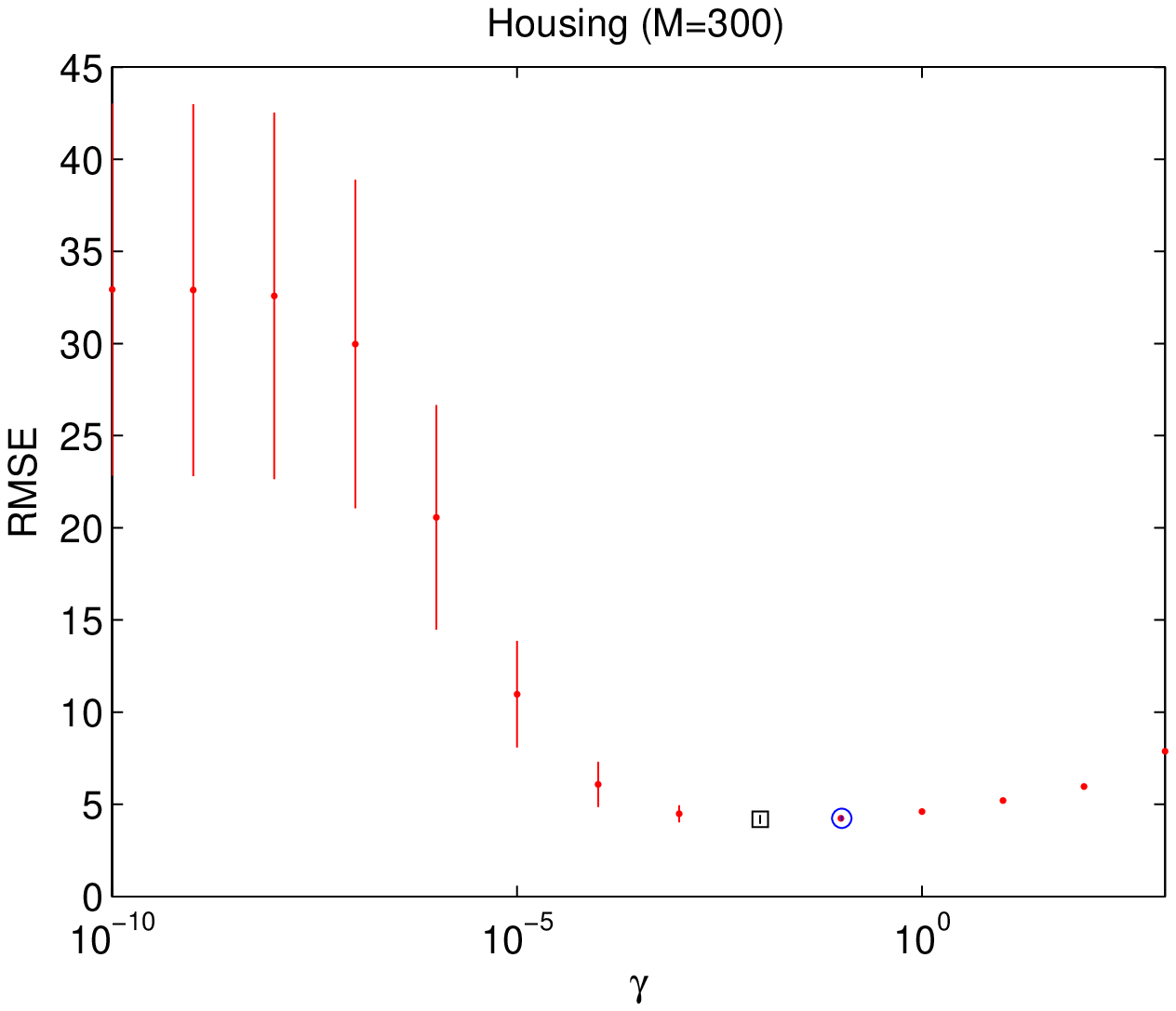}

\caption{ Test error trends for regression datasets as a function of the values of $\gamma$ over the selected 
cross-validation range (red dots): the cross-validation selected $\gamma$ is 
the black square; the proposed $\gamma$ from OCReP is the blue circle. 
 }
\label{fig:BoxPlotArray1}
\end{figure}

\begin{figure}
\centering
\includegraphics[width=0.49\textwidth]{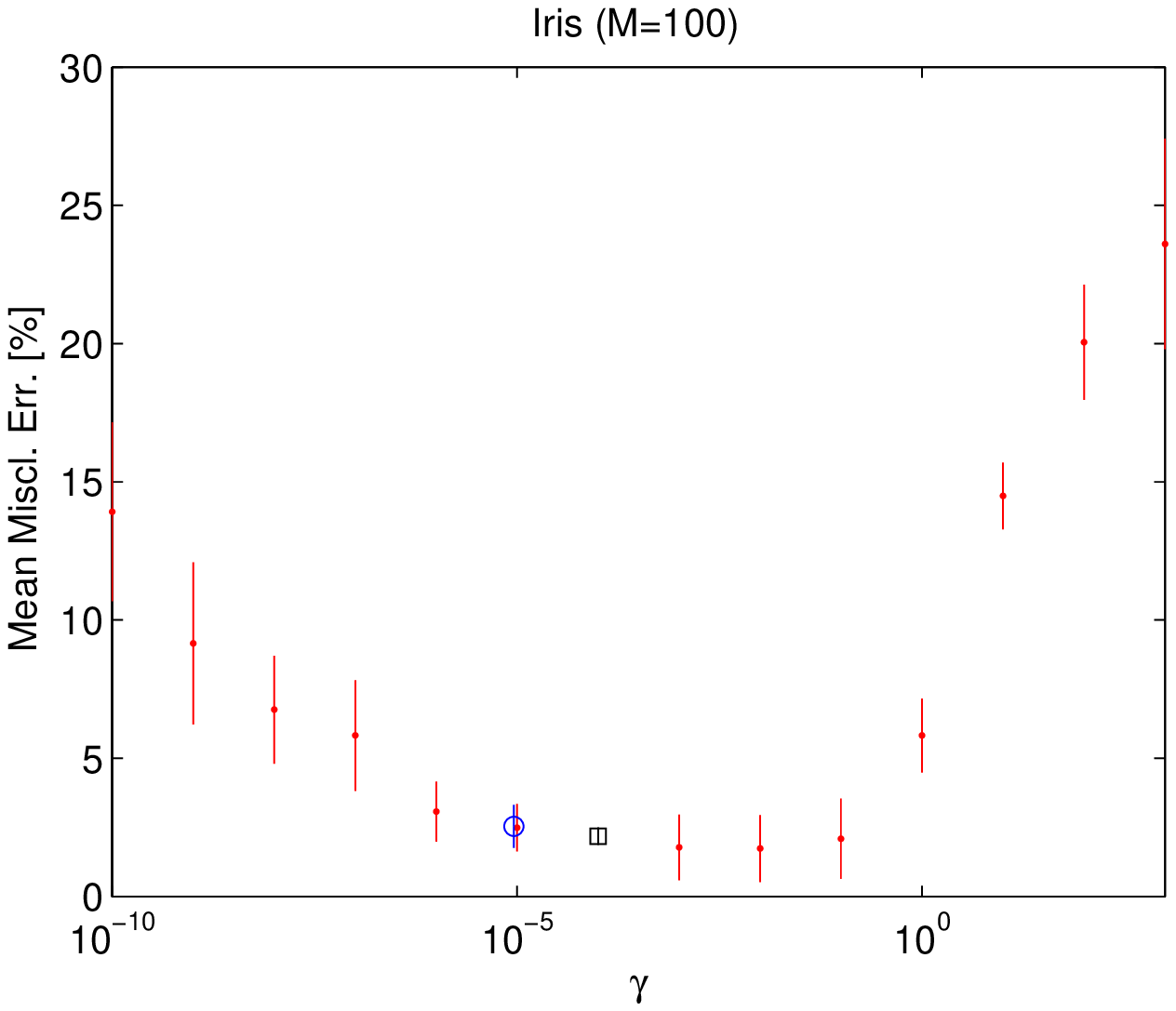}
\includegraphics[width=0.49\textwidth]{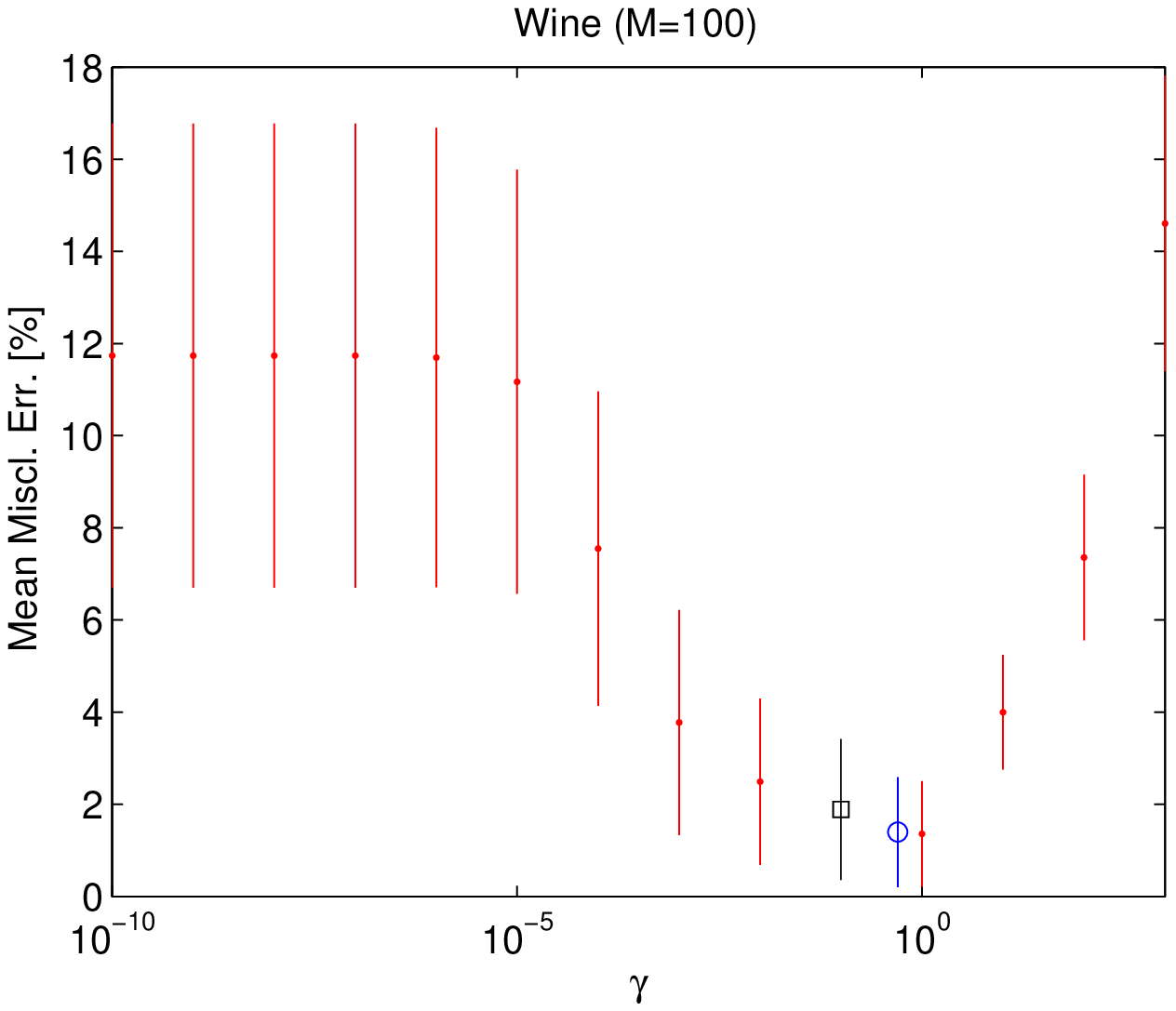}

\includegraphics[width=0.49\textwidth]{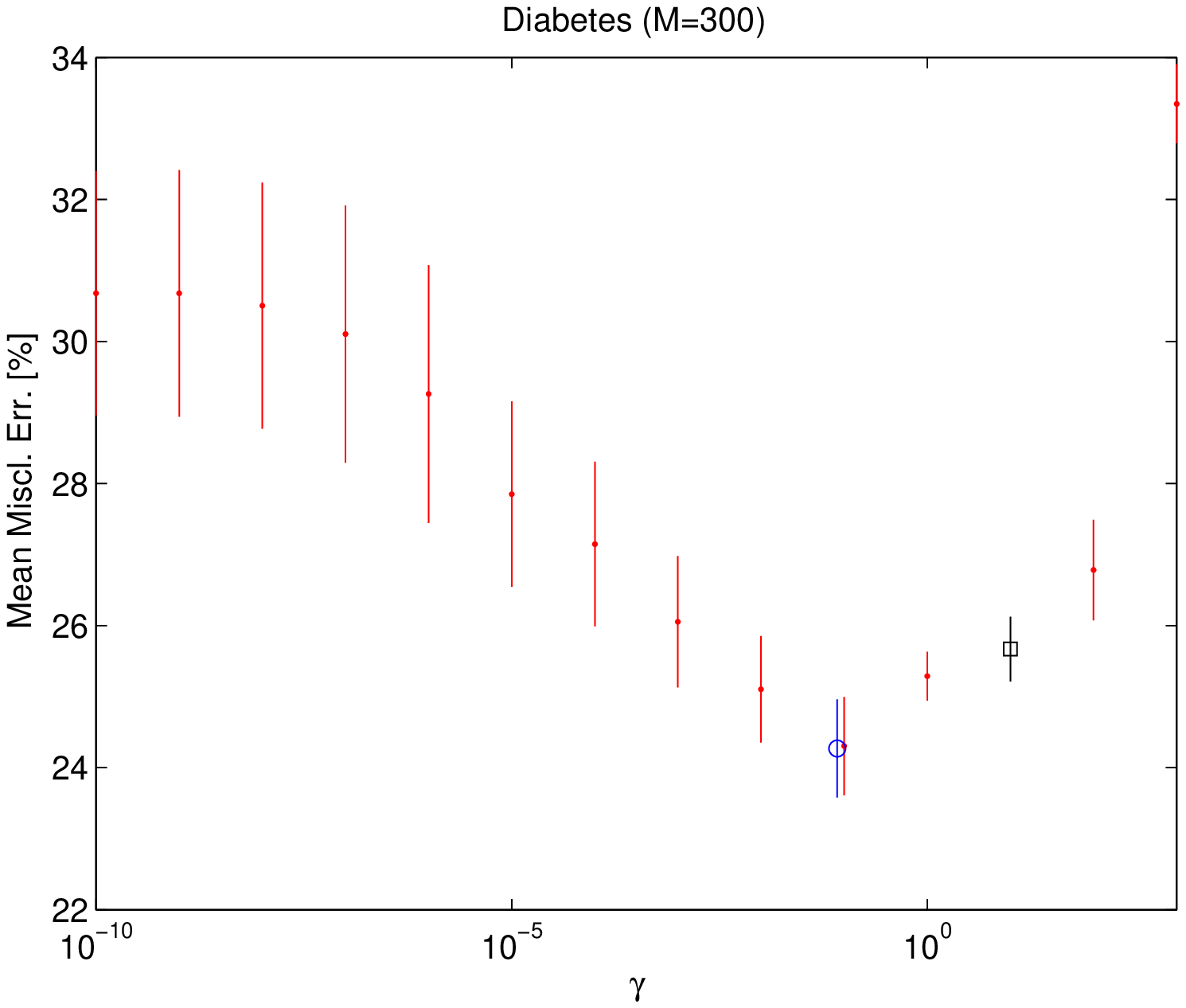}
\includegraphics[width=0.49\textwidth]{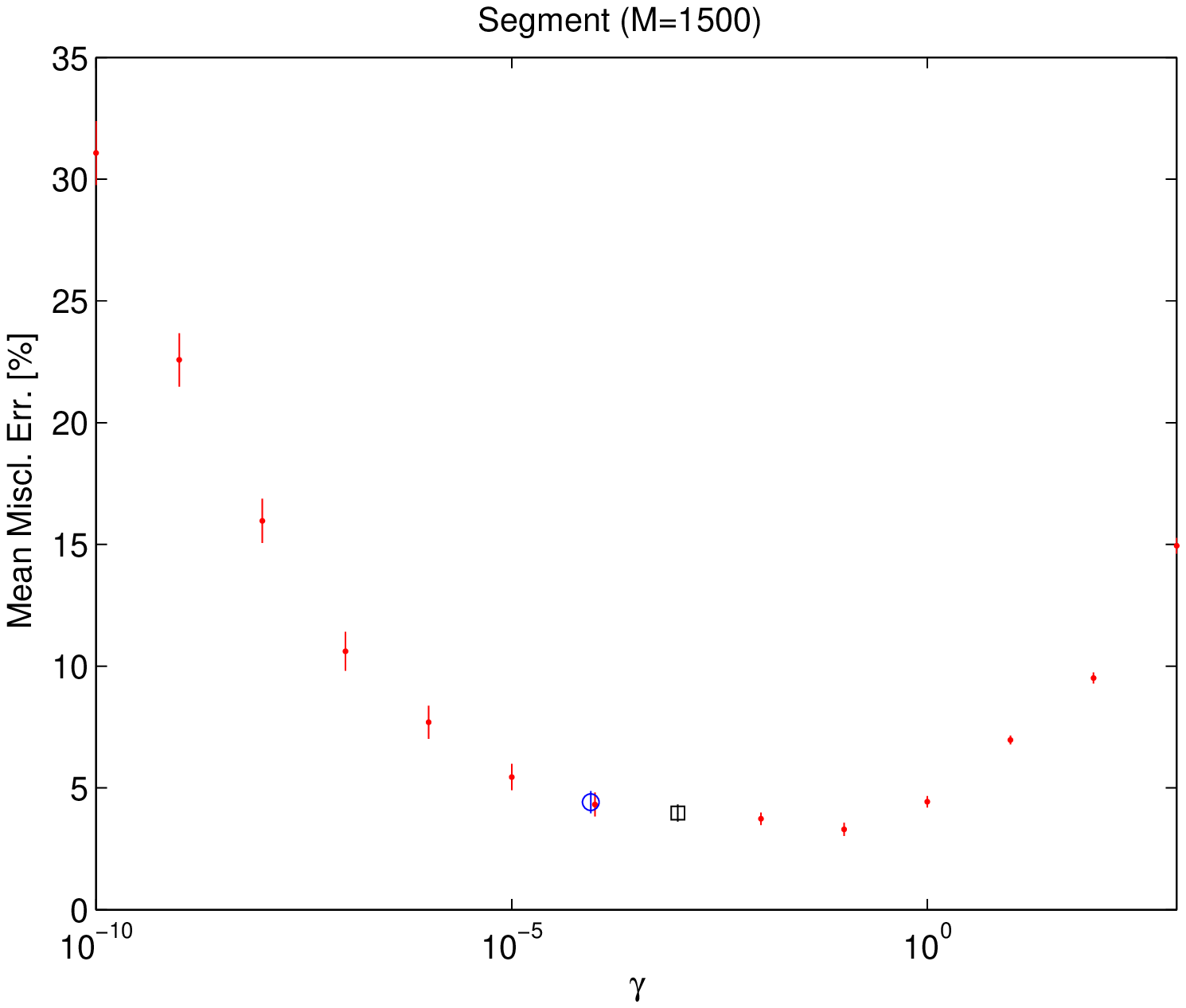}

\caption{ Test error trends for classification datasets as a function of the values of $\gamma$ over the selected 
cross-validation range (red dots): the cross-validation selected $\gamma$ is 
the black square; the proposed $\gamma$ from OCReP is the blue circle. 
 }
\label{fig:BoxPlotArray2}
\end{figure}

Figures \ref{fig:BoxPlotArray1} and \ref{fig:BoxPlotArray2} 
(respectively for regression and classification datasets) 
show average test errors as a function of the sampled values of $\gamma$ 
(red dots); the standard deviation is shown as an error bar. 
Our proposed optimal $\gamma$ is evidenced as a blue circle, whereas 
the value of $\gamma$ selected by cross-validation is shown as a black 
square. The results are in each case related to the highest number of neurons 
experimented. 

The horizontal axis has been zoomed in onto the region of interest, 
i.e. $[10^{-10}, \ 10^5]$.

It may be noted that the performance from OCReP and cross-validation 
are comparable, and also close to the experimental minimum. 
This may be interpreted as good predictivity for both algorithms.

Also, we remark that the error bars, i.e. experimental result dispersion, 
is large for small values of $\gamma$, consistently with expectations on 
ineffective regularization.

\begin{table}[ht]
\caption{Comparison of OCReP vs. cross-validation at fixed number of hidden 
neurons for small size datasets }
\begin{center}
\begin{tabular}{c|c|ccccc}
\multicolumn{4}{c}{ } & $\quad$ Iris $\quad$ & Wine & Machine Cpu\\ 
\hline
\multirow{8}{*}{$M$} & \multirow{4}{*}{$50$} & \multirow{2}{*}{OCReP} & Err. & ${\bf 1.51}$ & $2.98$ & $31.21$\\
                             &  &  & Std & $1.13$ & $1.75$ & $1.1$ \\
                             \cline{3-7}
                             &  & \multirow{2}{*}{cross-val.} & Err. & $2.13$ & $3.37$ & $31.1$\\
                             &  &  & Std & $0.77$ & $2.27$ & $1.02$\\[2mm]
                             \cline{2-7}
                             & \multirow{4}{*}{$100$} & \multirow{2}{*}{OCReP} & Err. & $2.53$ & ${\bf 1.39}$ & $34.13$\\
                             &  &  & Std & $0.77$ & $1.19$ & $01.68$\\
                             \cline{3-7}
                             &  & \multirow{2}{*}{cross-val.} & Err. & ${\bf 2.17}$ & $1.88$ & ${\bf 30.94}$ \\
                             &  &  & Std & $0.31$ & $1.88$ & $0.69$ \\ \hline
\end{tabular}
\end{center}
\label{tab:ocrepcvpiccoli}
\end{table}

\begin{table}[h]
\caption{Comparison of OCReP vs. cross-validation at fixed number of hidden 
neurons for large size datasets}
\begin{center}
\begin{tabular}{c|c|ccc}
\multicolumn{4}{c}{ } & $\quad$ Segment $\quad$\\ 
\hline
\multirow{8}{*}{$M$} & \multirow{4}{*}{$1000$} & \multirow{2}{*}{OCReP} & Err. & $2.53$ \\
                             &  &  & Std & $0.77$ \\
                             \cline{3-5}
                             &  & \multirow{2}{*}{cross-val.} & Err. & ${\bf 2.17}$\\
                             &  &  & Std & $0.31$  \\[2mm]
                                                          \cline{2-5}
                             & \multirow{4}{*}{$1500$} & \multirow{2}{*}{OCReP} & Err. & $4.41$ \\
                             &  &  & Std & $0.45$ \\
                             \cline{3-5}
                             &  & \multirow{2}{*}{cross-val.} & Err. & ${\bf 3.97}$\\
                             &  &  & Std & $0.35$  \\                             
                              \hline
\end{tabular}
\end{center}
\label{tab:ocrepcvgrandi}
\end{table}

\begin{table}[ht]
\caption{Comparison of OCReP vs. cross-validation at fixed number of hidden 
neurons for medium size datasets. For Delta Ailerons, average errors and 
standard deviations have to be multiplied by $10^{-4}$.} 
\begin{center}
\begin{tabular}{c|c|cccccc}
\multicolumn{4}{c}{ } & Abalone & Delta Ailerons & Housing & Diabetes\\ 
\hline
\multirow{16}{*}{$M$} & \multirow{4}{*}{$50$} & \multirow{2}{*}{OCReP} & Err. & $2.22$ & $1.64$ & $5.54$ & ${\bf 26.01}$\\
                             &  &  & Std & $0.16$ & $0.0051$ & $0.12$ & $0.604$\\
                             \cline{3-8}

                             &  & \multirow{2}{*}{cross-val.} & Err. & ${\bf2.13}$ & ${\bf1.59}$ & ${\bf4.79}$ & $26.79$\\
                             &  &  & Std & $0.017$ & $0.0073$ & $0.37$ & $0.814$\\[2mm]
                             \cline{2-8}
                             & \multirow{4}{*}{$100$} & \multirow{2}{*}{OCReP} & Err. & $2.15$ & $1.62$ & $5.17$ & $25.66$\\
                             &  &  & Std & $0.007$ & $0.004$ & $0.08$ & $0.608$\\
                             \cline{3-8}
                             &  & \multirow{2}{*}{cross-val.} & Err. & ${\bf 2.11}$ & ${\bf 1.58}$ & ${\bf 4.49} $ & $25.71$ \\
                             &  &  & Std & $0.006$ & $0.0036$ & $0.28$ & $0.608$ \\[2mm]
                             \cline{2-8}                                                          
                             & \multirow{4}{*}{$200$} & \multirow{2}{*}{OCReP} & Err. & $2.12$ & ${\bf 1.59}$ & $4.62$ & ${\bf 25.13}$\\
                             &  &  & Std & $0.003$ & $0.0031$ & $0.09$ & $0.445$\\
                             \cline{3-8}
                             &  & \multirow{2}{*}{cross-val.} & Err. & ${\bf 2.11} $ & $1.61$ & ${\bf 4.30}$ & $25.79$\\
                             &  &  & Std & $0.003$ & $0.0096$ & $0.27$ & $0.443$\\[2mm] 
                             \cline{2-8}                                                          
                             & \multirow{4}{*}{$300$} & \multirow{2}{*}{OCReP} & Err. & $2.113$ & ${\bf 1.58}$ & $4.24$ & ${\bf 24.26}$\\
                             &  &  & Std & $0.03$ & $0.0018$ & $0.13$ & $0.689$\\
                             \cline{3-8}
                             &  & \multirow{2}{*}{cross-val.} & Err. & $2.114$ & $1.60$ & $4.18$ & $25.66$ \\
                             &  &  & Std & $0.003$ & $0.0042$ & $0.23$ & $0.456$ \\                              
                             \hline
\end{tabular}
\end{center}
\label{tab:ocrepcvmedi}
\end{table}

The numerical results have been reported in Tab.~\ref{tab:ocrepcvpiccoli}, 
\ref{tab:ocrepcvgrandi} and \ref{tab:ocrepcvmedi} according to the 
grouping based on dimension and complexity of the datasets. 

For each dataset and selected number of hidden neurons $M $, the best test 
error is evidenced in bold, whenever the difference is statistically 
significant\footnote{The Student's t-test has been used for assessing 
the statistical significance through determination of the confidence 
intervals related to $99\%$ confidence level.}. 

Thus, for example, on Iris the best performance is achieved using 50 neurons by OCReP, and with 100 neurons by cross-validation. 
In some cases, e.g. Wine (50 neurons), there is no clear winner from 
statistical considerations, i.e. the best results are comparable, within 
the errors. 

From the above results it appears that cross-validation has better test error performance on 
a number of datasets slightly higher,  at fixed number of hidden neurons. 
However, it is important to evidence that the use of OCReP allows to save the hundreds of pseudoinversion steps required by cross-validation, wich is a crucial issue for practical implementation.

\subsection {OCReP performance assessment: variable number of hidden units}
\label{sec:OCReP_unreg}

In order to pursue the double aim of performance and hidden units optimization, a first interesting step is to give a look to the variation as a function of hidden layer dimension of error trends of unregularized models (i.e. models whose output weights are evaluated according to eq.(\ref{eq:5b})).

A context widely used among researchers using such techniques (see e.g. \citet{helmy,huang}) is to use input weights distributed according to a random uniform distribution in the interval $(-1, 1)$, and sigmoidal activation functions for hidden neurons: hereafter we name this framework Sigm-unreg.

\begin{figure}[h]
\centering

\includegraphics[width=0.49\textwidth]{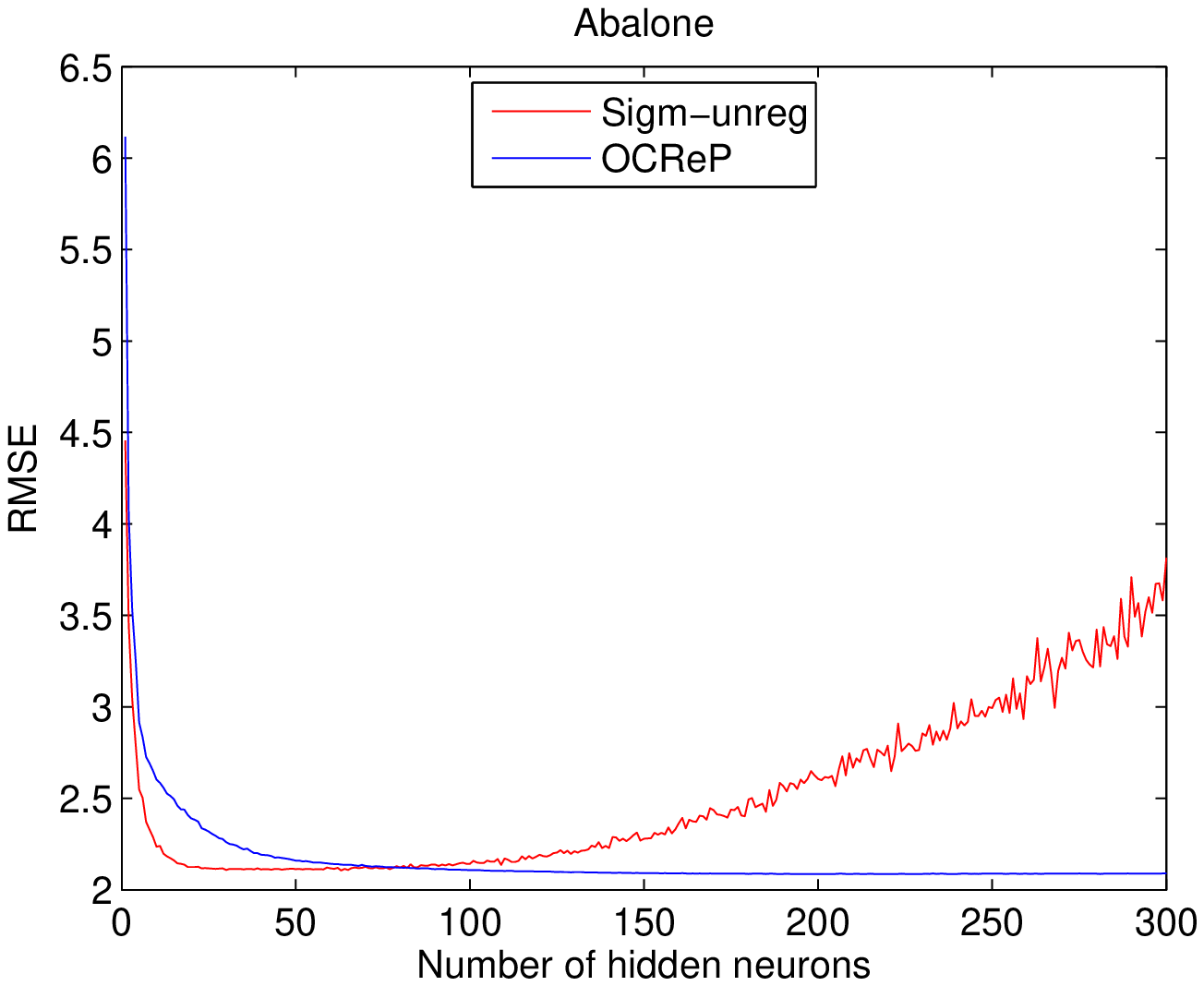}
\includegraphics[width=0.49\textwidth]{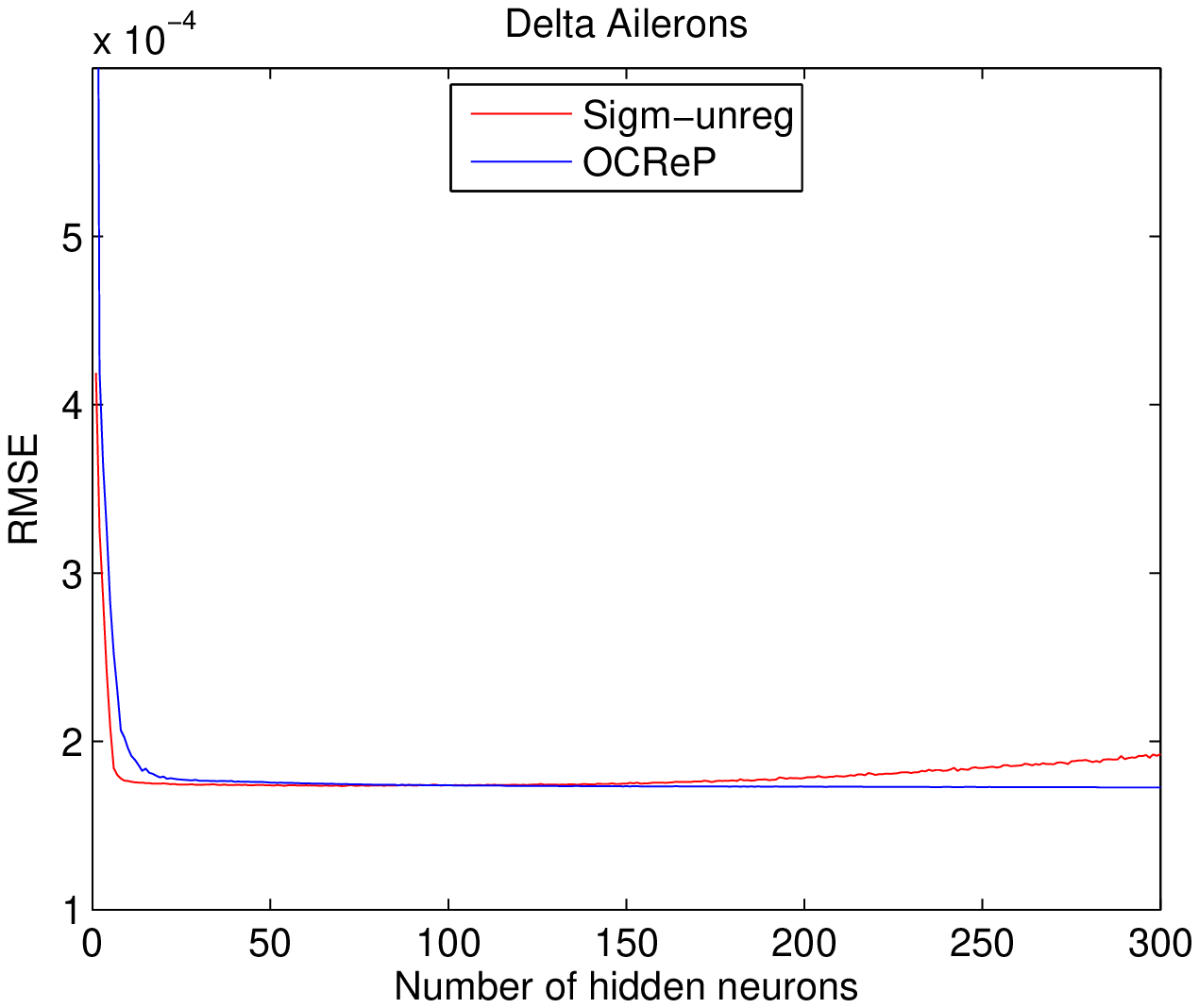}

\includegraphics[width=0.49\textwidth]{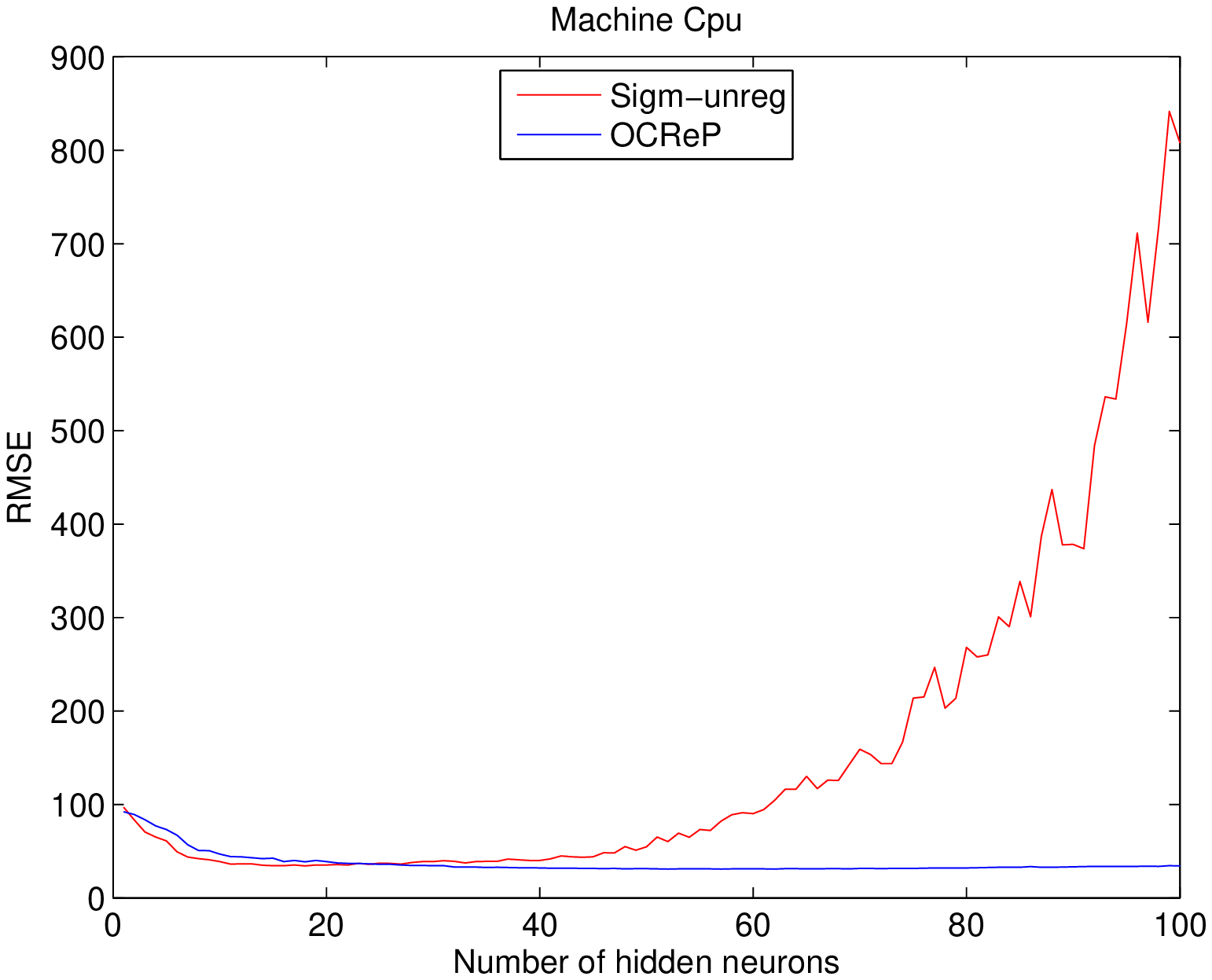}
\includegraphics[width=0.49\textwidth]{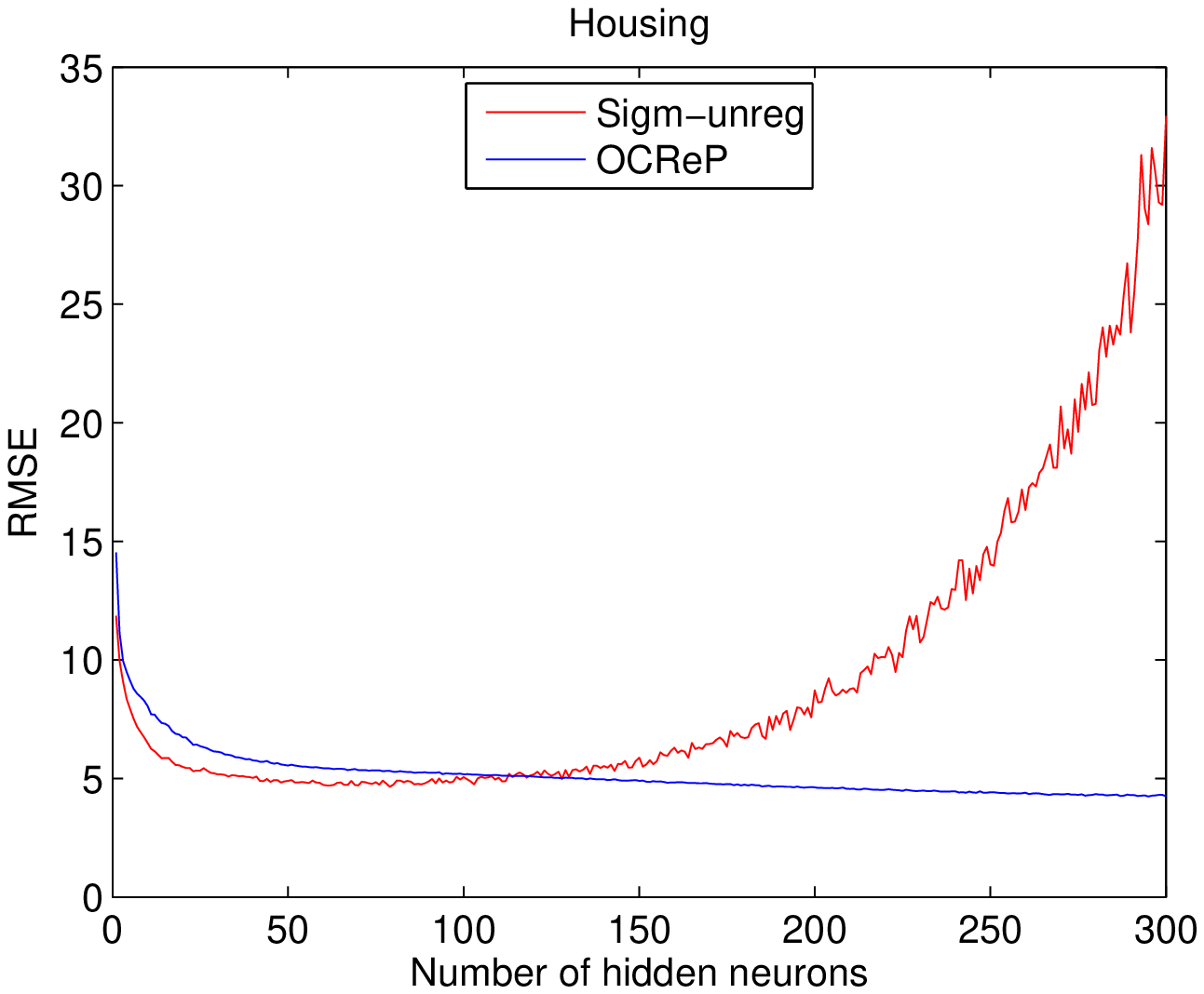}

\caption{Test error trends for regression datasets: OCReP vs. unregularized pseudoinversion. }
\label{fig:PlotArray1}
\end{figure} 

\begin{figure}
\centering

\includegraphics[width=0.49\textwidth]{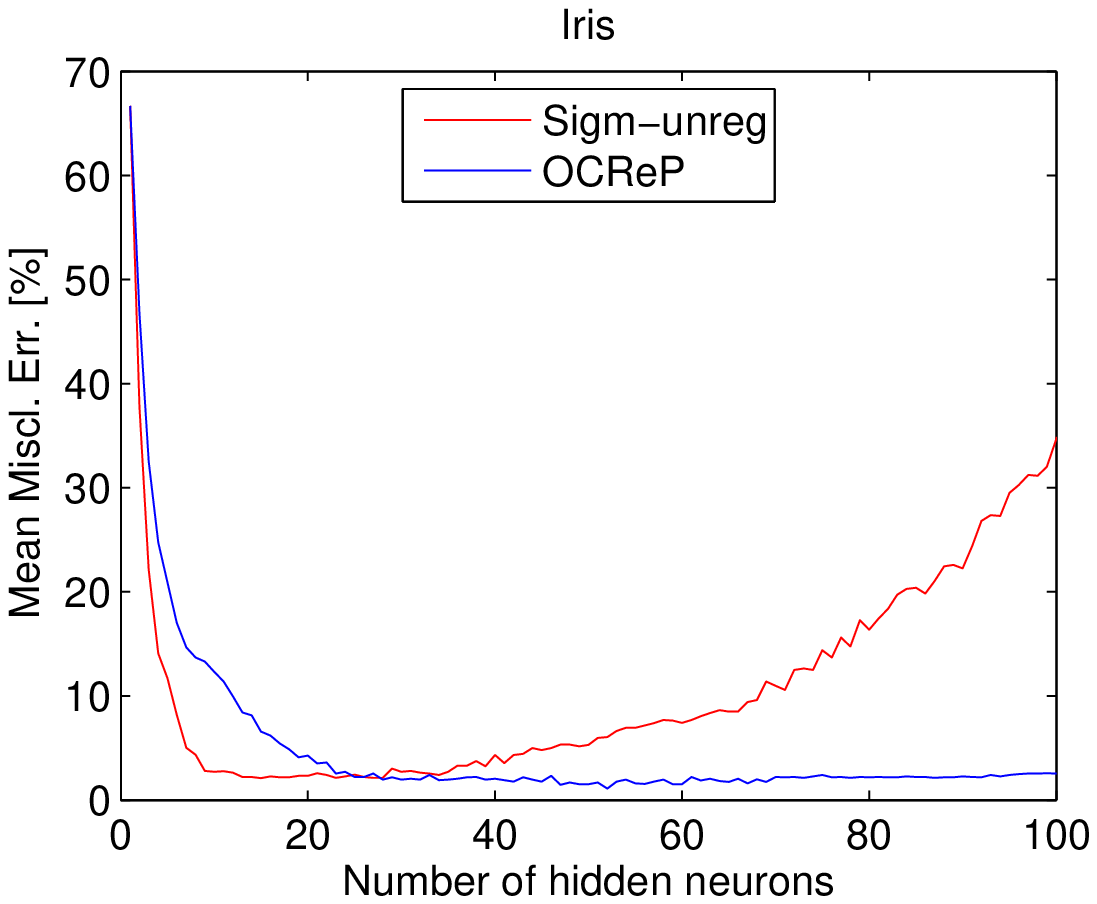}
\includegraphics[width=0.49\textwidth]{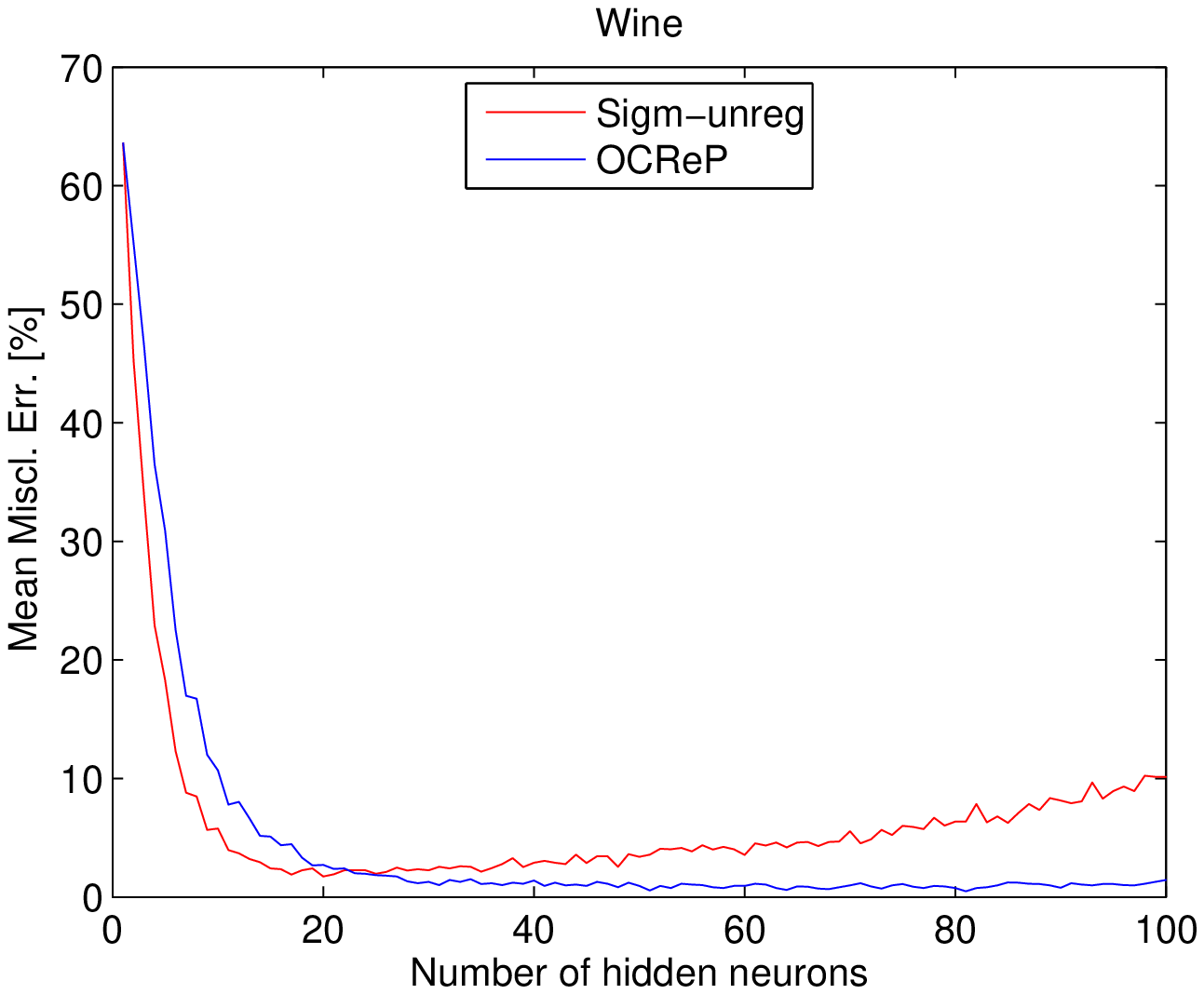}

\includegraphics[width=0.49\textwidth]{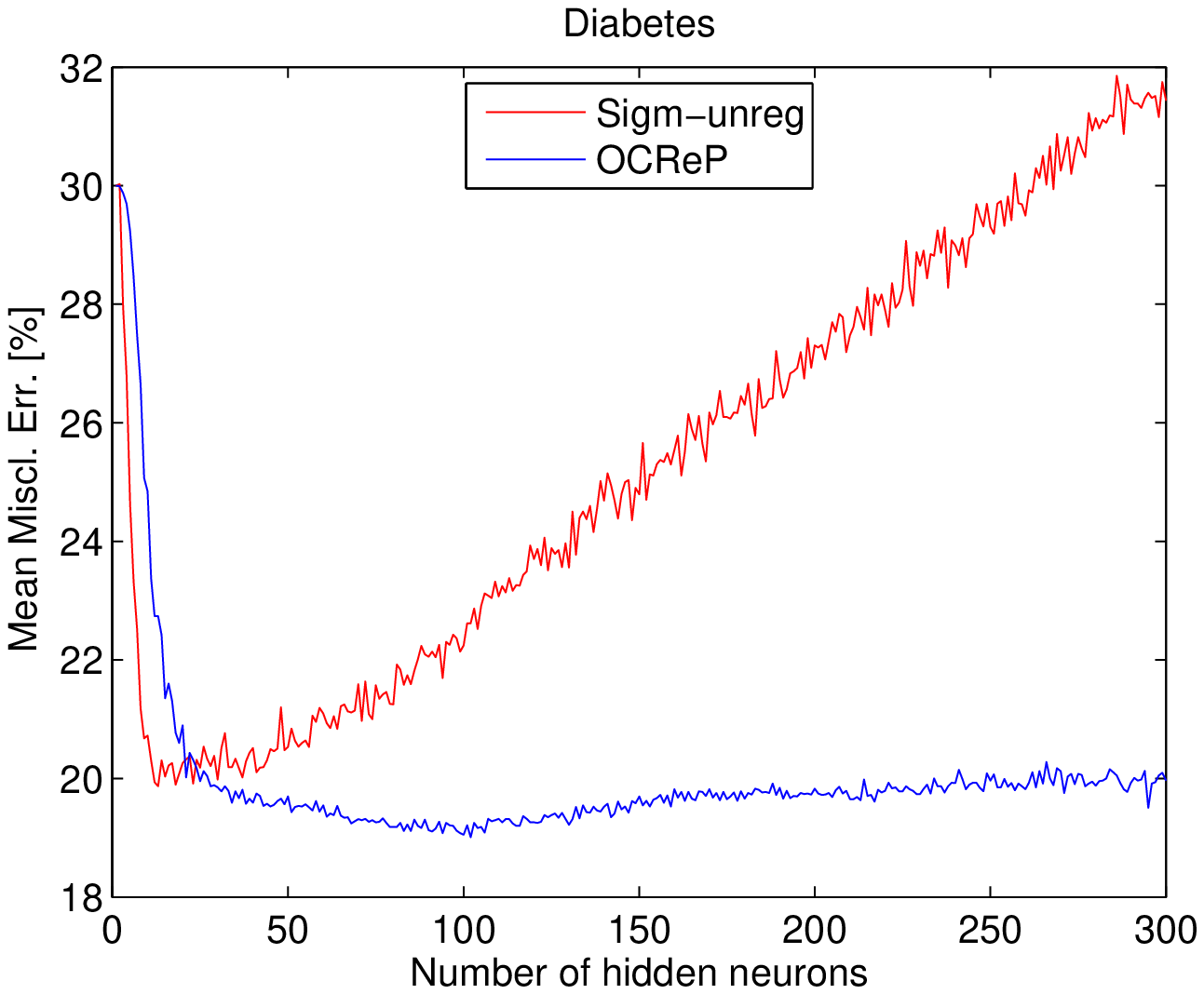}
\includegraphics[width=0.49\textwidth]{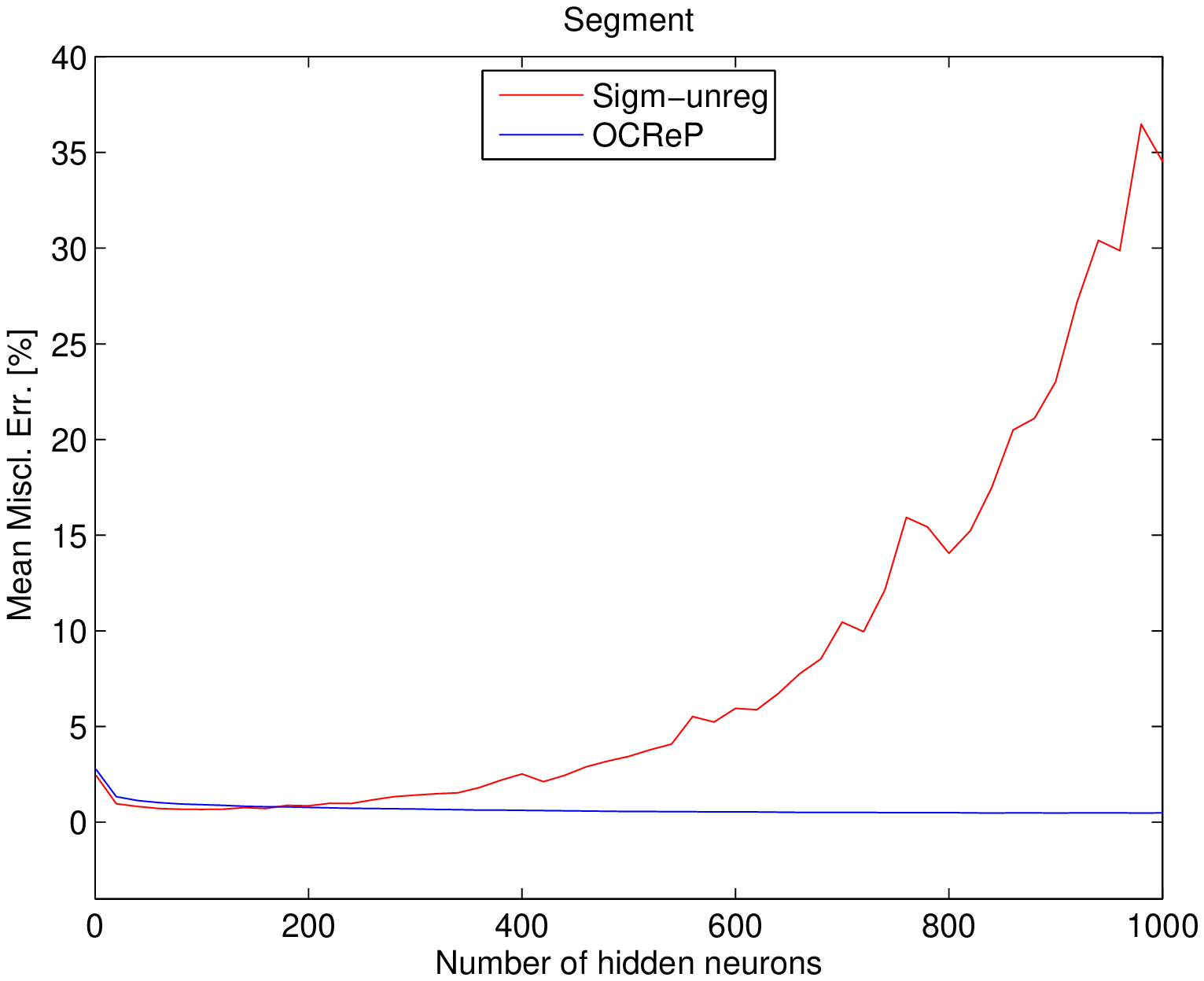}

\caption{Test error trends for classification datasets: OCReP vs. unregularized pseudoinversion. }
\label{fig:PlotArray2}
\end{figure}

Figures \ref{fig:PlotArray1} and \ref{fig:PlotArray2} show, respectively for regression and classification datasets, the average test error values, (over 50 different input weights selections) for both OCReP (blue line) and Sigm-unreg (red line) as a function of the number of hidden nodes, which is gradually increased by unity steps. 
In all cases, after an initial decrease the Sigm-unreg test error increases significantly.

On the contrary, the OCReP test error curves keep decreasing, albeit at 
slower and slower rate, thus showing also a good capability of overfitting control of the method.

We aim now at comparing the results obtained when the trade-off value of $ \gamma $ is searched by cross-validation, with the two different frameworks discussed so far, i.e. OCReP and Sigm-unreg. 

A 70\%/30\% split between training and test set is applied; we then perform 
a three-fold cross-validation for the selection of the number of hidden 
neurons $\bar{M}$ at which the minimum error is recorded in all cases. 
Test errors are again evaluated as the average of 50 different random choices 
of input weights. 

The numerical results of the simulation are presented in Tables 
\ref{tab:perf_regr} and \ref{tab:perf_class}, respectively for regression 
and classification tasks, with their standard deviations (Std) and $\bar{M}$. 

Best test errors are evidenced in bold, whenever the difference between OCReP and cross-validation is statistically 
significant.

We see that our proposed regularization technique provides, for regression datasets, performance comparable with the cross-validation option but always a better performance (with statistical significance at $99\%$ level) with respect to the unregularized case.

For classification datatsets in three cases out of four OCReP provides a 
better performance with respect to cross-validation, and always a better 
performance with respect to the Sigm-unreg case. 
In all such cases, the statistical significance is at the $99\%$ level. 

Also, in almost all cases smaller standard deviations are associated with 
the OCReP method, suggesting a lower sensitivity to initial input weights 
conditions.

\subsection { Additional considerations } 
\label{sec:final_cond} 

The proposed method OCReP presents in our opinion two features of interest:
on one side, its computational efficiency, and on the other side its 
optimal conditioning. 

Our goal of optimal analytic determination of the regularization parameter 
$\gamma $ results in a dramatic improvement in the computing requirements 
with respect to experimental tuning by search over a pre-defined large grid 
of $N_\gamma$ tentative values. 
In the latter case, for each choice of $\gamma $ over the selected range, 
at least a pseudoinversion is required for every output weight determination, thus increasing the computational load by a factor $N_\gamma$.

Besides, our method is designed explicitly for optimal conditioning. 
In our simulations, we verify that the goal is fulfilled by evaluating 
average condition numbers of hidden layer output matrices. 
The statistics is performed over 50 different configurations of input 
weights and a fixed number of hidden units, namely the largest used 
in section \ref{sec:OCReP_CV} for each dataset.
The results are summarised in Tables \ref{tab:ncondmediograndi} and 
\ref{tab:ncondpiccoliegrande}, respectively for regression and 
classification datasets. 
On the first row of each table, we list the ratio of average condition numbers 
of matrices $H^{reg}$, and $H^+$, associated respectively to OCReP and 
Sigm-unreg, i.e. regularized and unregularized approaches. 
On the second row, we list the ratio of average condition numbers of matrices $H^{reg}$ and $H^{CV}$, thus comparing our regularization approach with the 
more conventional one, the latter using cross-validation. 

Not surprisingly, our regularization method provides a significant improvement 
on conditioning with respect to the unregularized approach, as evidenced by 
ratio values much smaller than unity. 
Besides, OCReP also provides better conditioned matrices than those derived 
by selection of $\gamma$ through cross-validation, since the corresponding 
condition numbers are systematically smaller in the former case, sometimes 
up to an order of magnitude.

\begin{table}[ht]
\caption{Hidden layer optimization for regression tasks. For Delta Ailerons, average errors and standard deviations have to be multiplied by $10^{-4}$.}
\begin{center}
\begin{tabular}{lcccc}
 & Abalone & Housing & Delta Ailerons & Machine Cpu \\
\cline{1-5}
 & \multicolumn{4}{c}{OCReP} \\
\cline{1-5}
Err. 		& $2.12$ & $4.25$ & $1.58$ & ${\bf 31.22}$ \\
Std. 		& $0.32$ & $0.13$ & $0.0048$ & $0.78$ \\
$\bar{M}$ & $178$ & $255$ & $298$ & $63$ \\
\hline
& \multicolumn{4}{c}{Cross-validation} \\
\hline
Err.      & ${\bf 2.11}$ & $4.19$ & $1.58$ & $31.51$ \\
Std. 		& $0.0097$ & $0.25$ & $0.0036$ & $1.25$ \\
$\bar{M}$ & $110$ & $250$ & $93$ & $70$ \\
\hline
& \multicolumn{4}{c}{Sigm-unreg} \\
\hline
Err.      & $2.14$ & $4.73$ & $1.62$ & $34.44$ \\
Std. 		& $0.014$ & $0.20$ & $0.57$ & $2.89$ \\
$\bar{M}$ & $31$ & $76$ & $74$ & $15$ \\
\hline
\end{tabular}
\end{center}
\label{tab:perf_regr}
\end{table}

\begin{table}[ht]
\caption{Hidden layer optimization for classification tasks. }
\begin{center}
\begin{tabular}{lcccc}
 			& Iris & Wine & Diabetes & Segment \\
\hline
 & \multicolumn{4}{c}{OCReP} \\
\cline{1-5}
Err.  		& ${\bf 1.6}$ & ${\bf 1.73}$ & $25.53$ & ${\bf 2.50}$ \\
Std. 		& $1.10$ & $1.25$ & $0.51$ & $0.32$ \\
$\bar{M}$ & $67$ & $91$ & $291$ & $760$ \\
\hline
& \multicolumn{4}{c}{Cross-validation} \\
\hline
Err.  & $2.12$ & $2.10$ & $25.2$ & $2.65$ \\
Std. 		& $1.26$ & $2.27$ & $1.29$ & $0.38$ \\
$\bar{M}$ & $14$ & $137$ & $25$ & $620$ \\
\hline
& \multicolumn{4}{c}{Sigm-unreg} \\
\hline
Err.  		& $2.31$ & $3.20$ & $25.92$ & $4.45$ \\
Std. 		& $1.48$ & $2.09$ & $1.12$ & $0.47$ \\
$\bar{M}$ & $67$ & $91$ & $291$ & $760$ \\
\hline
\end{tabular}
\end{center}
\label{tab:perf_class}
\end{table}

\begin{table}[ht]
\caption{Condition number comparison for regression datasets}
\begin{center}
\begin{tabular}{lcccc}
    & Abalone & Housing & Delta Ailerons & Machine Cpu \\ \hline
    $\mu(H^{reg}) /  \mu(H^+)$    & $0.0002$ & $0.0008$ & $0.00007$ & $0.0001$ \\ 
    $\mu(H^{reg}) / \mu(H^{CV})$ & $0.8$   & $0.3$ & $0.3$ & $0.1$ \\
    \hline
\end{tabular}
\end{center}
\label{tab:ncondmediograndi}
\end{table}

\begin{table}[ht]
\caption{Condition number comparison for classification datasets }
\begin{center}
\begin{tabular}{lcccc}
    & Iris & Wine & Diabetes & Segment\\ \hline
    $\mu(H^{reg}) /  \mu(H^+)$   & $0.00002$ & $0.005$ & $0.0007$ &  $0.000005$ \\ 
    $\mu(H^{reg}) / \mu(H^{CV})$ & $0.2$   & $0.4$ & $0.1$ & $0.2$ \\
    \hline
\end{tabular}
\end{center}
\label{tab:ncondpiccoliegrande}
\end{table}


\section{Comparison with other approaches}
\label{comparison}

Since the literature provides a host of different recipes 
for either the choice of the regularization parameter, or 
the actual regularization algorithm, hereafter we focus on 
a couple of specific frameworks.

\begin{table}
\caption{GCV results at fixed number of hidden neurons for small datasets}
\begin{center}
\begin{tabular}{c|c|cccc}
\multicolumn{3}{c}{ } & $\quad$ Iris $\quad$ & Wine & Machine Cpu \\
\hline
\multirow{4}{*}{$M$} & \multirow{2}{*}{$50$}  & Err. & $2.47$ & $3.66$ & $33.03$ \\
                             & &  Std & $1.06$ & $2.42$ & $1.27$ \\
                             & \multirow{2}{*}{$100$} & Err. & $3.06$ & $3.77$ & $36.06$ \\
                             &  & Std & $1.08$ & $2.44$ & $1.13$ \\                       
                              \hline
\end{tabular}
\end{center}
\label{tab:golubpiccoli}
\end{table}

\begin{table}
\caption{GCV results at fixed number of hidden neurons for large size datasets}
\begin{center}
\begin{tabular}{c|c|ccc}
\multicolumn{4}{c}{ } & $\quad$ Segment $\quad$\\ 
\hline
\multirow{4}{*}{$M$}  & \multirow{2}{*}{$1000$} & \multirow{2}{*}{} & Err. & $11.39$ \\
                             &  &  & Std & $0.75$ \\
                             \cline{2-5}
                             & \multirow{2}{*}{$1500$} & \multirow{2}{*}{} & Err. & $14.72$ \\
                             &  &  & Std & $0.803$ \\                           
                              \hline
\end{tabular}
\end{center}
\label{tab:golubsegment}
\end{table}

\begin{table}
\caption{GCV results at fixed number of hidden neurons for medium size datasets. 
For Delta Ailerons, average errors and standard deviations have to be multiplied by $10^{-4}$.}
\begin{center}
\begin{tabular}{c|c|ccccccc}
\multicolumn{4}{c}{ } & Abalone & Housing & Delta Ailerons & Diabetes\\ 
\hline
\multirow{8}{*}{$M$} & \multirow{2}{*}{$50$} & \multirow{2}{*}{} & Err. & $ \bf {2.13}$ & $ \bf {4.89}$ & ${\bf 1.60}$ & ${\bf 25.2}$ \\
                             &  &  & Std & $0.017$ & $0.45$ & $0.0103$ & $1.22$ \\
                             \cline{2-8}
                             & \multirow{2}{*}{$100$} & \multirow{2}{*}{} & Err. & $2.15$ & $5.05$ & $1.63$ & $26.66$ \\
                             &  &  & Std & $0.021$ & $0.70$ & $0.0297$ & $1.39$ \\
                             \cline{2-8}
							 & \multirow{2}{*}{$200$} & \multirow{2}{*}{} & Err. & $2.32$ & $6.78$ & $1.74$ & $27.73$ \\
                             &  &  & Std & $0.10$ & $2.35$ & $0.0892$ & $1.27$ \\
                             \cline{2-8}
                             & \multirow{2}{*}{$300$} & \multirow{2}{*}{} & Err. & $2.98$ & $8.07$ & $2.20$ & $27.14$ \\
                            &  &  & Std & $0.42$ & $2.89$ & $0.4054$ & $1.15$ \\                         
                              \hline
\end{tabular}
\end{center}
\label{tab:golubmediograndi}
\end{table}

\subsection{ Other choices of regularization parameter }

Among the approaches mentioned in section \ref{OLSRR}, 
we primary select the technique of generalised cross-validation (GCV) 
from \citep{golubwahba}, described by eqs.~(\ref{eq:Golub1}) 
and (\ref{eq:Golub2}), for comparison with our method. The main motivation for our choice is its independence on the estimate of the error variance $\sigma^2$, which is a characteristic shared with our 
case. 
For each dataset, we select the same fixed numbers of hidden units 
as in section \ref{sec:OCReP_CV}: then for each case eq.~(\ref{eq:Golub1}) 
is minimized 
over the set of 50 values of  $\gamma$ $[10^{-25}, 10^{-24} \cdots 10^{25}]$ 
and for 50 different configurations of input weights. 
 
We evaluate the mean and standard deviation 
of the corresponding regularized test error, reported in 
Tables~\ref{tab:golubpiccoli}, \ref{tab:golubsegment} and 
\ref{tab:golubmediograndi}. 
We also remind that the tabulated error ``Err" is either the average RMSE 
for regression tasks, or the average misclassification rate for 
classification tasks; ``Std" is the corresponding standard deviation. 
The performance comparison is based on statistical significance at $99\%$ 
level. 

Whenever GCV provides test error values statistically better than OCReP
(listed in Tab.~\ref{tab:ocrepcvpiccoli}, \ref{tab:ocrepcvgrandi} 
and \ref{tab:ocrepcvmedi}), they are marked in bold. 

We remark that in all cases listed in Tab.~\ref{tab:ocrepcvpiccoli} 
and \ref{tab:ocrepcvgrandi} OCReP provides statistically better results 
than GCV. 
The situation of medium size datasets evidences a somewhat mixed 
behaviour: with 50 hidden neurons, GCV wins; with 100 neurons, 
for three out of four datasets (i.e. Abalone, Housing and Diabetes) 
the performance is statistically comparable. 
In all other cases of Tab.~\ref{tab:ocrepcvmedi} OCReP again provides 
better statistical results than GCV.

We make two other comparisons, using the ridge estimates described in eq.(13) and eq.(9) of \citep{DorugadeKashid}, and proposed respectively by \citep{kibria} and \citep{HoerlKennard70}: 

\begin{equation}
\gamma_K = {1 \over p} \sum_1^p {\hat \sigma^2 \over \hat \alpha_i^2 } \ , 
\label{eq:kibria_est}
\end{equation}

\begin{equation}
\gamma_{HK} = {\hat \sigma^2 \over \hat \alpha_{max}^2 } \ . 
\label{eq:hoerl_est}
\end{equation}

Our experimentation is made only for regression datasets because the 
theoretical background of  \citep{DorugadeKashid}, and of most of  other works referred in section \ref{OLSRR}, directly applies to the case 
in which the quantity $Y$ in eq.(\ref{eq:ORL}) is a one column matrix. In our formulation $Y$ is the desired target $T$ and it is  a one-column matrix only for regression tasks.

For each dataset we applied both methods described by eq. \ref{eq:kibria_est} and \ref{eq:hoerl_est}; we select the same fixed numbers of hidden units as in section \ref{sec:OCReP_CV} and perform 50 experiments with different configuration of input weights.

Each step of pseudoinversion is regularized for each method with the corresponding $\gamma$ value.
We evaluate the mean and standard deviation 
of the regularized test errors, reported respectively in 
Tables~\ref{tab:kibriaregress} and \ref{tab:hoerlregress}. 

Whenever the methods provide test error values statistically better than 
OCReP (listed in Tab.~\ref{tab:ocrepcvpiccoli} and \ref{tab:ocrepcvmedi}), 
they are marked in bold. 

We remark that the method by Kibria obtains a better performance in two cases over sixteen, while OCReP in 12 cases over sixteen.
Besides, the method by Hoerl and Kennard obtains a better performance 
in three cases over sixteen, while OCReP in eight cases over sixteen. 
For both methods, better performance is achieved only for the case of 
$M =50$ neurons. 

It may be noted that with respect to processing requirements OCReP has 
clear advantages, since it requires only a SVD step for each determination 
of $\gamma$, while the above two methods require full spectral decomposition 
and an additional matrix inversion. 

\subsection{Alternative regularization methods }

A first comparison can be done with the work by Huang et al. \citep{Huang2}, 
whose technique Extreme Learning Machine (ELM) uses a cost 
parameter $C$ that can be considered as related to the inverse 
of our regularization parameter $ \gamma $. 
As authors state, in order to achieve good generalization performance, $C$ 
needs to be chosen appropriately. 
They do this by trying 50 different values of this parameter: 
$ [2^{-24}, 2^{-23}, \cdots 2^{24}, 2^{25}]$.

A fair comparison can be done on our classification datasets, using 
their number of hidden neurons, i.e. 1000. 
Our optimal choice of $\gamma$ allows to obtain a better performance on all 
datasets (with statistical significance assessed at the same confidence level that previous experiments).

\begin{table}
\caption{ Kibria estimate of ridge parameter: results at fixed number of hidden neurons for regression datasets.
For Delta Ailerons, average errors and standard deviations have to be multiplied by $10^{-4}$.}
\begin{center}
\begin{tabular}{c|c|ccccccc}
\multicolumn{4}{c}{ } & Abalone & Housing & Delta Ailerons & Machine Cpu\\ 
\hline
\multirow{8}{*}{$M$} & \multirow{2}{*}{$50$} & \multirow{2}{*}{} & Err. & $2.32$ & $5.72$ & ${\bf 1.63}$ & ${\bf 34.28}$ \\
                             &  &  & Std & $0.37$ & $0.84$ & $0.027$ & $4.67$ \\
                             \cline{2-8}
                             & \multirow{2}{*}{$100$} & \multirow{2}{*}{} & Err. & $2.38$ & $5.45$ & $1.64$ & $32.40$ \\
                             &  &  & Std & $0.90$ & $0.86$ & $0.08$ & $3.72$ \\
                             \cline{2-8}
							 & \multirow{2}{*}{$200$} & \multirow{2}{*}{} & Err. & $2.20$ & $5.31$ & $1.65$ &  \\
                             &  &  & Std & $0.13$ & $0.76$ & $0.15$ &  \\
                             \cline{2-8}
                             & \multirow{2}{*}{$300$} & \multirow{2}{*}{} & Err. & $2.34$ & $5.46$ & $1.62$ &  \\
                            &  &  & Std & $1.01$ & $1.60$ & $0.035$ & \\                         
                              \hline
\end{tabular}
\end{center}
\label{tab:kibriaregress}
\end{table}

\begin{table}
\caption{H-K estimate of ridge parameter: results at fixed number of hidden neurons for regression datasets.
For Delta Ailerons, average errors and standard deviations have to be multiplied by $10^{-4}$.}
\begin{center}
\begin{tabular}{c|c|ccccccc}
\multicolumn{4}{c}{ } & Abalone & Housing & Delta Ailerons & Machine Cpu\\ 
\hline
\multirow{8}{*}{$M$} & \multirow{2}{*}{$50$} & \multirow{2}{*}{} & Err. & ${\bf 2.13}$ & ${\bf 4.87}$ & ${\bf 1.60}$ & $34.28$ \\
                             &  &  & Std & $0.016$ & $0.44$ & $0.01$ & $2.37$ \\
                             \cline{2-8}
                             & \multirow{2}{*}{$100$} & \multirow{2}{*}{} & Err. & $2.14$ & $4.98$ & $1.62$ & $37.39$ \\
                             &  &  & Std & $0.90$ & $0.67$ & $0.029$ & $3.18$ \\
                             \cline{2-8}
							 & \multirow{2}{*}{$200$} & \multirow{2}{*}{} & Err. & $2.33$ & $8.101$ & $1.73$ &  \\
                             &  &  & Std & $0.10$ & $2.83$ & $0.08$ &  \\
                             \cline{2-8}
                             & \multirow{2}{*}{$300$} & \multirow{2}{*}{} & Err. & $2.95$ & $29.06$ & $2.21$ &  \\
                            &  &  & Std & $0.41$ & $9.26$ & $0.41$ & \\                         
                              \hline
\end{tabular}
\end{center}
\label{tab:hoerlregress}
\end{table}

\begin{table}[ht]
\caption{Comparison between OCReP and ELM }
\begin{center}
\begin{tabular}{c|ccccc}
\multicolumn{2}{c}{ } & $\quad$ Iris $\quad$ & Wine & Diabetes & Segment\\ 
\hline
\multirow{2}{*}{OCReP} & Err. & ${\bf 2.22}$ & ${\bf 1.28}$ & ${\bf 21.06}$ & ${\bf 3.40}$\\
					   & Std. & $0.21$ & $0.88$ & $0.65$ & $0.25$\\
                             \hline
\multirow{2}{*}{ELM} & Err. & $2.4$ & $1.53$ & $22.05$ & $3.93$\\
					   & Std  & $2.29$ & $1.81$ & $2.18$ & $0.69$\\
 \hline
\end{tabular}
\end{center}
\label{tab:huang}
\end{table}

Deng et al. \citep{deng} propose a Regularized Extreme Learning Machine (hereafter, RELM) in wich the regularization parameter is selected according to a similar criterion among 100 values: $ [2^{-50}, 2^{-49}, \cdots 2^{50}]$. 
Because their performance is optimized with respect to the number of hidden neurons, for the sake of comparison we use OCReP values from table \ref{tab:perf_class}.
We obtain a statistically significant better performance on dataset Segment, while for Diabetes the method RELM performs better (see table \ref{tab:deng}).

\begin{table}[ht]
\caption{Comparison between OCReP and RELM }
\begin{center}
\begin{tabular}{c|ccc}
\multicolumn{2}{c}{ } &  Diabetes & Segment\\ 
\hline
\multirow{3}{*}{OCReP} & Err. & $25.53$ & ${\bf 2.50}$ \\
					   & Std. & $0.51$ & $0.32$ \\
					   & $\bar{M}$ & $291$ & $760$ \\
                             \hline
\multirow{3}{*}{RELM} & Err. & ${\bf 21.81}$ & $4.49$ \\
					   & Std.  & $2.55$ & $0.0074$ \\
					  & $\bar{M}$ & $15$ & $200$ \\
 \hline
\end{tabular}
\end{center}
\label{tab:deng}
\end{table}

Comparing our results on the common regression datasets 
with the alternative method TROP-ELM 
proposed by Miche et al. \citep{Miche}, 
we note that OCReP achieves always lower RMSE values \footnote{In that work, 
performance and related statistics are expressed in terms of MSE; 
we only derived the corresponding RMSE for comparison with our results. } 
(with statistical significance), as can be seen from table \ref{tab:miche}. 

Besides, in our opinion our method is simpler, in the sense that 
it uses a single step of regularization rather than two.

\begin{table}[ht]
\caption{Comparison between OCReP and TROP-ELM. For Delta Ailerons, average errors and standard deviations have to be multiplied by $10^{-4}$. }
\begin{center}
\begin{tabular}{c|ccccc}
\multicolumn{2}{c}{ } & Abalone & Delta Ailerons & Machine Cpu & Housing \\ 
\hline
\multirow{3}{*}{OCReP} & Err. & ${\bf 2.12}$ & ${\bf 1.58}$ & ${\bf 31.22}$ & ${\bf 4.25}$\\
					   & Std. & $0.32$ & $0.0048$ & $0.78$ & $0.13$\\
                              & $\bar{M}$ & $178$ & $298$ & $63$ & $255$\\ \hline
\multirow{3}{*}{TROP-ELM} & Err. & $2.19$ & $1.64$ & $264.03$ & $34.35$\\
                                        & $\bar{M}$ & $42$ & $80$ & $28$ & $59$\\
 \hline
\end{tabular}
\end{center}
\label{tab:miche}
\end{table}

In \citep{Martinez}, an algorithm is proposed for pruning ELM networks 
by using regularized regression methods: the crucial step of regularization parameter determination 
is solved by creating $K$ different models, each one based on a different value of this parameter, 
among which the best one is selected using a Bayesian information criterion. 
Authors state that a typical value for $K$ is 100, thus an heavy computational load is required, and the method is focused on regression tasks.

\section{Conclusions}
In the context of regularization techniques for single hidden 
layer neural networks trained by pseudoinversion, we provide 
an optimal value of the regularization parameter $\gamma$ by 
analytic derivation. 
This is achieved by defining a convenient regularized 
matricial formulation in the framework of Singular Value 
Decomposition, in which the regularization parameter is derived 
under the constraint of condition number minimization. 
The OCReP method has been tested on UCI datasets for both regression 
and classification tasks. 
For all cases, regularization implemented using the analytically derived 
$\gamma$ is proven to be very effective in terms of predictivity, as 
evidenced by comparison 
with implementations of other approaches from the literature, including 
cross-validation. 
OCReP avoids hundreds of pseudoinversions usually needed by most other 
methods, i.e. it is quite computationally attractive. 

\section*{Acknowledgements}
The activity has been partially carried on in the context of the 
Visiting Professor Program of the Gruppo Nazionale per il Calcolo 
Scientifico (GNCS) of the Italian Istituto Nazionale di Alta 
Matematica (INdAM). 
This work has been partially supported by ASI contracts 
(Gaia Mission - The Italian Participation to DPAC) I/058/10/0-1 
and 2014-025-R.0. 


\bibliographystyle{natbib}
\bibliography{Cancelliere}

\end{document}